\documentclass[journal]{IEEEtran}
\usepackage{amsmath,amsfonts}
\usepackage{algorithmic}
\usepackage{algorithm}
\usepackage{array}
\usepackage[caption=false,font=normalsize,labelfont=sf,textfont=sf]{subfig}
\usepackage{textcomp}
\usepackage{stfloats}
\usepackage{url}
\usepackage{verbatim}
\usepackage{graphicx}
\usepackage{cite}
\usepackage{hyperref}
\usepackage{xcolor}         
\usepackage{float}
\usepackage{caption} 
\usepackage{siunitx}
\usepackage{multirow}
\usepackage{makecell}
\usepackage{booktabs}
\usepackage{url}
\usepackage{academicons}
\definecolor{orcidlogocol}{HTML}{A6CE39}
\begin{document}

\title{HPE-CogVLM: Advancing Vision Language Models with a Head Pose Grounding Task}

\author{Yu Tian, Tianqi Shao, Tsukasa Demizu, Xuyang Wu, Hsin-Tai Wu
\thanks{Yu Tian, Tianqi Shao, Tsukasa Demizu, and Hsin-Tai Wu are with Docomo Innovations, Inc., Sunnyvale, US (e-mail: yu.tian@docomoinnovations.com, jshao@docomoinnovations.com, tsukasa.demizu@docomoinnovations.com, hwu@docomoinnovations.com).}
\thanks{Xuyang Wu is with Department of Computer Science and Engineering, Santa Clara University, Santa Clara, US (e-mail: elviswu0306@gmail.com).}
\thanks{Xuyang Wu and Hsin-Tai Wu are the corresponding author.}
\thanks{© 2026 IEEE. This is the author's version of an article accepted for publication in IEEE Transactions on Circuits and Systems for Video Technology. The final version is available at https://doi.org/10.1109/TCSVT.2026.3675940}}

\maketitle

\begin{abstract}

Head pose estimation (HPE) requires a sophisticated understanding of 3D spatial relationships to generate precise yaw, pitch, and roll angles. Previous HPE models, primarily CNN-based, rely on cropped close-up human head images as inputs and often lack robustness in real-world scenario. Vision Language Models (VLMs) can analyze entire images while focusing on specific objects through their attention mechanisms. In this paper, we propose a novel framework to improve the HPE accuracy by leveraging the object detection grounding capability of a VLM, referred to as CogVLM. We empirically find that directly LoRA fine-tuning of this VLM for the HPE task fails to achieve desirable HPE accuracy, while some model merging methods can improve accuracy but frequently produce blended invalid response formats, struggling to handle both object detection and HPE tasks simultaneously. To integrate HPE capability into CogVLM effectively, we develop a novel LoRA layer-based model merging method. This merging approach applies a high cosine similarity threshold and a “winner-takes-all” layer selection strategy, aligning attention to the HPE task while preserving original object detection knowledge. It successfully resolves issues with blended invalid response formats and improves accuracy. Results show that our HPE-CogVLM achieves a 31.5\% reduction in Mean Absolute Error over the current state-of-the-art CNN model, 6DRepNet, in cross-dataset evaluation. Furthermore, HPE-CogVLM outperforms both directly LoRA fine-tuned and task arithmetic-based merged VLMs across all HPE metrics.

\end{abstract}


\section{Introduction}
\noindent \IEEEPARstart{N}{owadays}, the head pose estimation (HPE) techniques have been widely studied in various fields such as attention estimation~\cite{fischer2018rt, gaze360_2019, cheng2018appearance}, face recognition~\cite{7780892,8255649,5959981, 7299081}, customer behavior analysis~\cite{6655785,wu2016head}, driver assistance systems~\cite{5443483,vora2018driver, hu2021temporal, hu2020robust} and human-robot interaction~\cite{app11125366}. Unlike general object pose estimation~\cite{DBLP:journals/tcsv/CaoZFZLX24, 10204028, DBLP:journals/tcsv/FengXLLW24, 10378090, 10378477, 10363348, DBLP:journals/tcsv/LiuSLZFW22}, which emphasizes an object’s overall structure or motion, HPE demands higher accuracy and granularity to meet the requirements of its critical applications. This task involves predicting the Euler angles (yaw, pitch, and roll) of human heads from images or videos. Recent research efforts on some CNN-based models like 6DRepNet~\cite{Hempel_2022}, HopeNet~\cite{DBLP:journals/corr/abs-1710-00925} and WHENet~\cite{zhou2020whenet} have made significant advancements in HPE.

\begin{figure*}[t]
  \centering
  
  \begingroup
  \setlength{\tabcolsep}{2pt}  
  \renewcommand{\arraystretch}{1.0}
  \begin{tabular}{@{}ccccc@{}}
    \includegraphics[height=3.0cm,keepaspectratio]{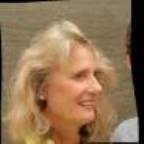} &
    \includegraphics[height=3.0cm,keepaspectratio]{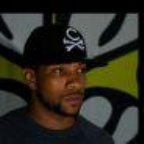} &
    \includegraphics[height=3.0cm,keepaspectratio]{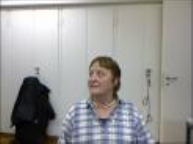} & \includegraphics[height=3.0cm,keepaspectratio]{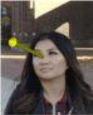} &
    \includegraphics[height=3.0cm,keepaspectratio]{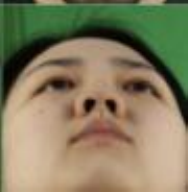} \\
    \small (a) 300W-LP & \small (b) AFLW2000 & \small (c) BIWI & \small (d) Gaze360 & \small (e) ETH-XGaze \\[-2pt]
    
     \\
  \end{tabular}
  \endgroup

  \begingroup
  \setlength{\tabcolsep}{2pt}
  \begin{tabular}{@{}ccc@{}}
  \includegraphics[height=3.0cm,keepaspectratio]{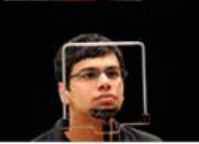} &
    \includegraphics[height=3.0cm,keepaspectratio]{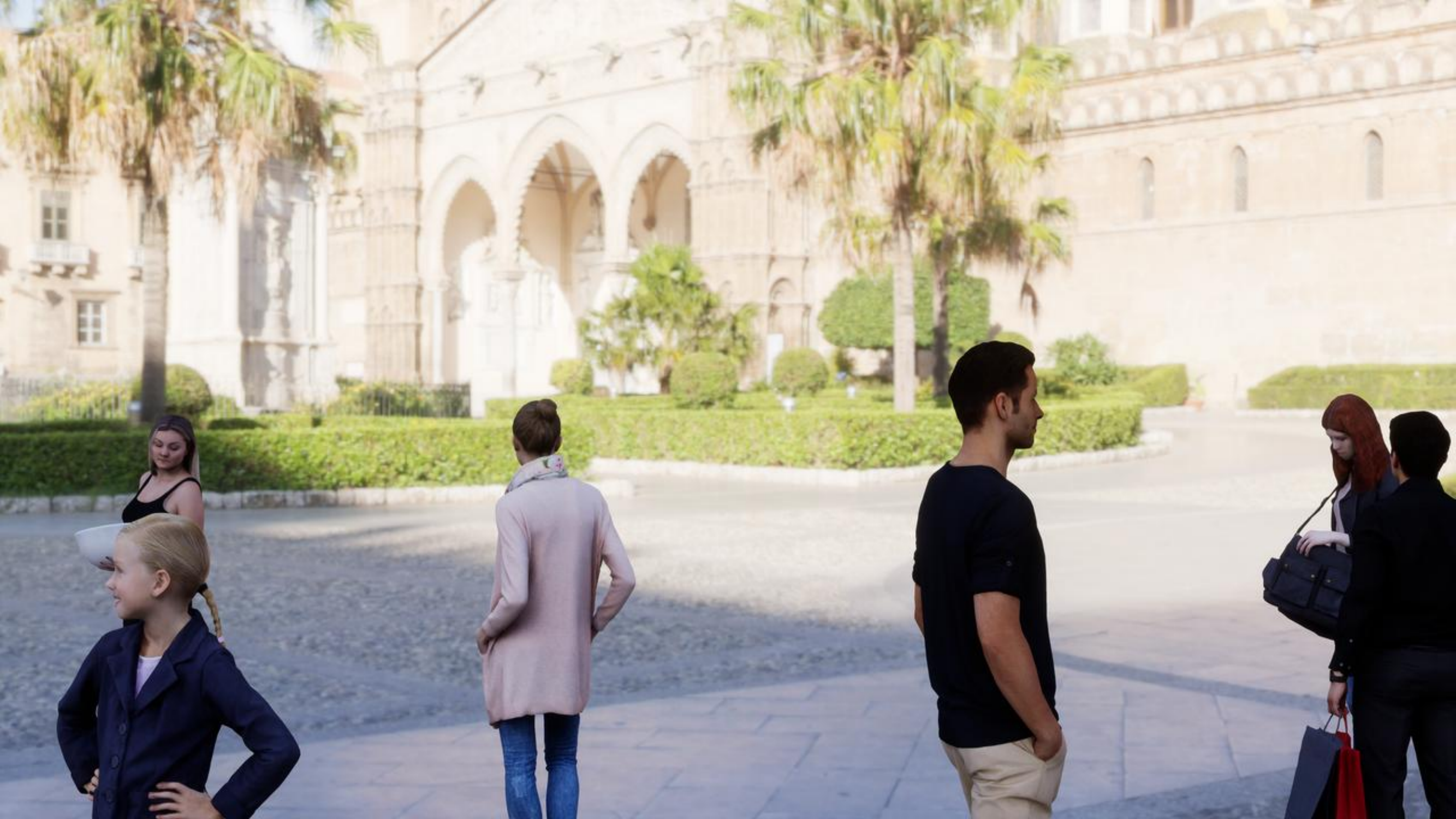} &
    \includegraphics[height=3.0cm,keepaspectratio]{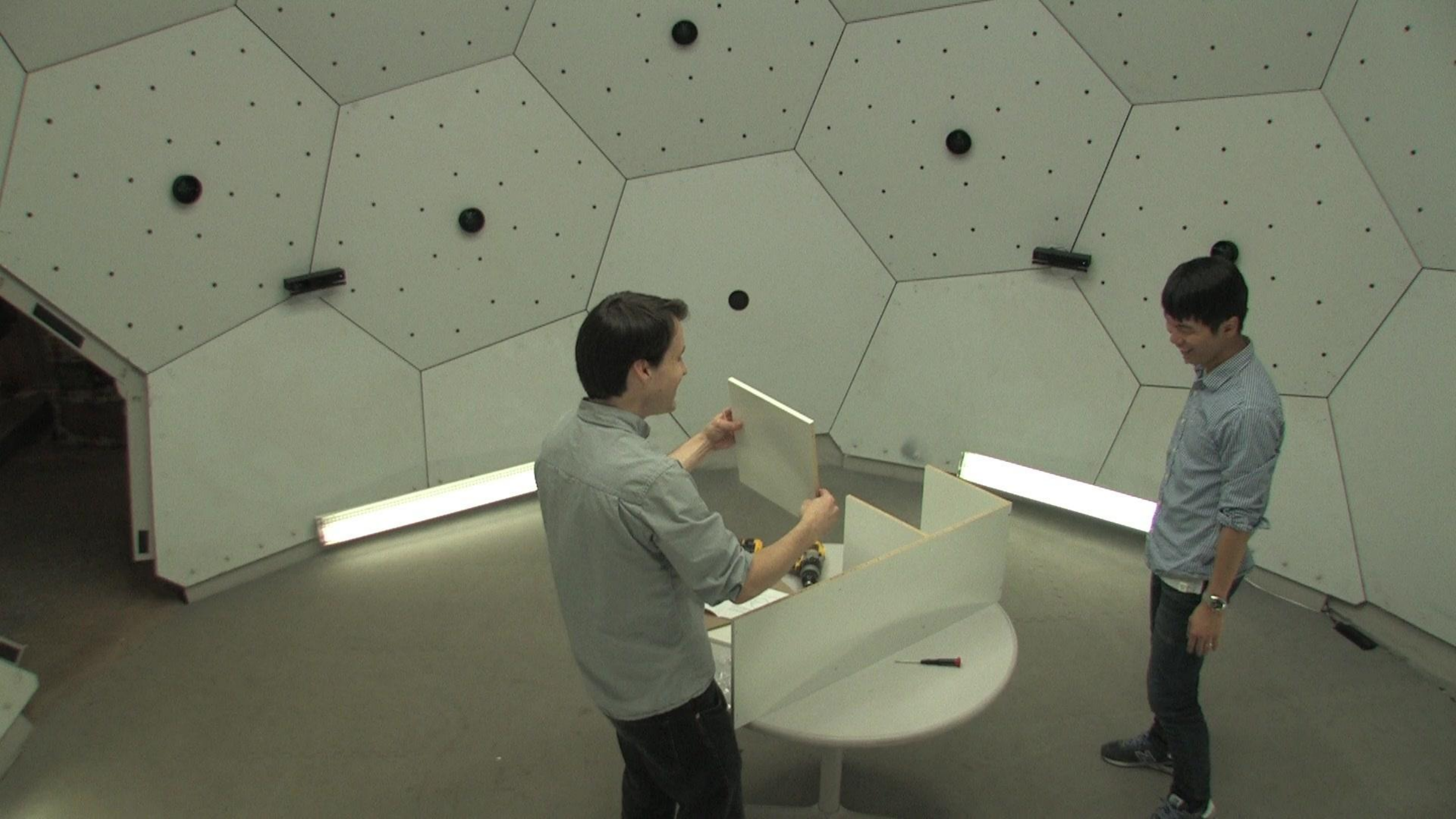} \\
    \small (f) Columbia Gaze & \small (g) Agora & \small (h) CMU Panoptic \\
  \end{tabular}
  \endgroup

  \caption{\textbf{Comparison of datasets for HPE input training data.}
  (a–f) Traditional datasets (300W-LP, AFLW2000, BIWI, Gaze360, ETH-XGaze, Columbia Gaze) mainly provide cropped head/face images with limited pose ranges and simplified backgrounds.
  (g–h) Our selected datasets (Agora, CMU Panoptic) offer full-scene, multi-person settings with richer variability and extreme/back-facing orientations, enabling more robust training for real-world scenarios.}
  \label{fig:hpe_comp}
\end{figure*}

\textit{Limitations of existing datasets.} Despite the recent surge of interest in HPE, real-world application still faces several challenges, particularly due to limitations in available datasets. As illustrated in Fig.~\ref{fig:hpe_comp}(a–c), most established HPE benchmarks, such as 300W-LP~\cite{Zhu_2019}, AFLW2000~\cite{Zhu_2019}, and BIWI~\cite{6553713}, primarily contain close-up images of single heads with yaw angles restricted to $[-99^\circ, 99^\circ]$, rather than the full $[-180^\circ, 180^\circ]$ range. 300W-LP further suffers from synthetic expansion artifacts that sometimes produce twisted or unrealistic facial appearances, while AFLW2000 is extremely limited in scale with only 2000 cropped samples. BIWI, although widely used, contains about 15,000 images captured from only 20 subjects (6 females and 14 males) in controlled indoor environments, which restricts both subject diversity and contextual variability. Other gaze-oriented datasets in Fig.~\ref{fig:hpe_comp}(d–f), such as ETH-XGaze~\cite{zhang2020ethxgazelargescaledataset}, Gaze360~\cite{gaze360_2019}, and Columbia Gaze~\cite{CAVE_0324}, cover a wider range of head orientations but still rely on close-up face or eye patches for training. As a result, these datasets focus narrowly on cropped frontal heads, reducing background diversity and often leading to overfitting.  

\textit{From cropped heads to full scenes.} These limitations highlight the need for robust HPE in real-world scenarios. In driver monitoring, retail analytics, and human–robot interaction, head orientation must be estimated even when facial features are distant, or unavailable, requiring methods that handle extreme poses and cluttered multi-person environments. For instance, detecting when a driver turns backward is critical for assessing attention, even without visible eye cues. To overcome these limitations, we leverage the Agora dataset~\cite{patel2021agora}, which provides dense full-pose coverage, multi-person configurations, and SMPL-X annotations~\cite{pavlakos2019expressive}, along with the CMU Panoptic dataset~\cite{joo2016panoptic}, which captures real-world full-scene multi-person interactions, as shown in Fig.~\ref{fig:hpe_comp}(g–h). In addition, the two datasets together encompass about 400 distinct persons, providing more than 60k head instances for training and 30k for testing as shown in Table \ref{datasets}. The DirectMHP model~\cite{zhou2023directmhp} attempts one-shot HPE prediction on these datasets, but struggles to maintain stability and to balance bounding box (BBox) detection task with accurate HPE task. These limitations highlight the need for a more flexible framework that can jointly localize heads and predict their orientations in unconstrained settings.

\begin{figure}[t!]
\small
\centering
\begin{minipage}{\linewidth}
\centering
\includegraphics[width=0.9\linewidth]{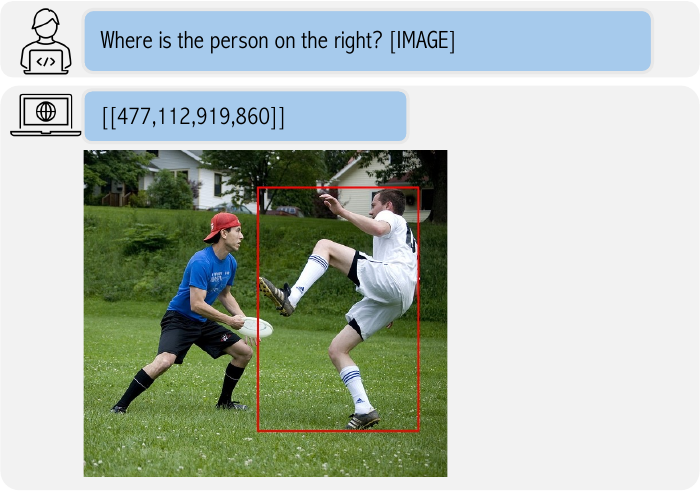}
\centering\mbox{\small (a) Example of CogVLM Grounding.}
\end{minipage}
\begin{minipage}{\linewidth}
\centering
\includegraphics[width=0.9\linewidth]{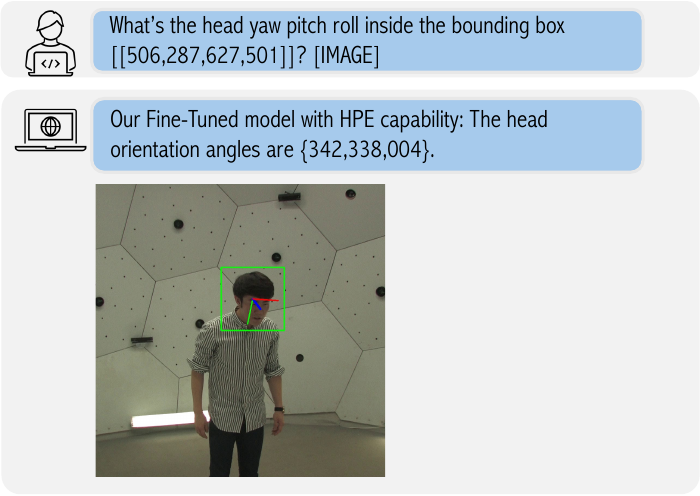}
\centering\mbox{\small (b) Example of HPE-CogVLM.}
\end{minipage}
\caption{Examples of CogVLM and HPE-CogVLM. (a) shows an example of CogVLM grounding capability, which demonstrates the original grounding CogVLM's ability to identify objects based on prompts, a foundational skill useful for HPE task. (b) displays a visualization of head orientation predicted by our HPE-CogVLM from the CMU Panoptic dataset, using Euler angles. The head pose labels are depicted with pitch (red axis), roll (green axis), and yaw (blue axis) angles, each indicated in their respective directions.
}
\label{fig:abl_rankagg_shot}
\end{figure}

The appearance of Large Language Models (LLMs) has made substantial advancements in a wide range of applications, significantly enhancing our daily lives by offering sophisticated assistance across various tasks. Recently, Vision Language Models (VLMs) have attracted significant attention for their capability to process multimodal information~\cite{openai2024gpt4, geminiteam2024gemini, liu2023visual, wang2024cogvlm}. By combining image and video understanding with language processing, VLMs can effectively perform complex tasks, such as visual question answering~\cite{ openai2024gpt4,agrawal2016vqa, liu2023visual} and visual grounding~\cite{wang2024cogvlm,yu2016modeling}. In this paper, we leverage the attention mechanism of VLMs to enhance the robustness and accuracy of the HPE task. The VLMs empower the model to process the entire image while focusing specifically on the head region, eliminating the need to crop out backgrounds. This capability reduces the risk of overfitting to a limited set of visual features and allows the model to leverage context from the full scene. As a result, it enhances the robustness of tasks that traditional CNN-based methods often struggle to address. We validate this approach by integrating HPE functionality into the grounding model of CogVLM~\cite{wang2024cogvlm}. The grounding CogVLM's capabilities include caption grounding, referring expression generation, referring expression comprehension and grounded visual question answering~\cite{wang2024cogvlm}. All of these functionalities involve in the description of object localization in the BBox format of [[$x_0$, $y_0$, $x_1$, $y_1$]] as shown in Figure \ref{fig:abl_rankagg_shot}(a). 
HPE is a natural extension of BBox grounding: while BBoxes provide 2D spatial localization, HPE additionally requires regressing 3D orientation (yaw, pitch, roll) of the localized head. Building on CogVLM’s grounding ability, we extend it to structured numerical outputs for HPE, unifying spatial localization and orientation analysis within a single VLM framework.

Despite its advantages, incorporating the HPE task into the grounding CogVLM introduces several challenges. First, VLM tasks such as image description, visual reasoning, and visual perception usually contain answering questions with natural language responses. In contrast, our HPE task requires the VLM to produce precise numerical Euler angles. Although the grounding CogVLM can predict BBoxes, indicating its ability to produce numerical responses, the HPE task is significantly more complicated. HPE requires predicting the human head's orientation in terms of yaw, pitch, and roll angles, which involves interpreting 3D orientation from 2D images. This introduces additional dimensions of depth and angular perspective not required in the basic BBox detection task. Therefore, it raises the challenge of whether the grounding model can provide HPE answers with much higher accuracy. Secondly, catastrophic forgetting~\cite{scialom2022finetuned, huang2024mitigating,luo2024empirical} poses a significant challenge in fine-tuning LLMs. The catastrophic forgetting problem is a phenomenon that LLMs tend to forget previously learned information when acquiring new data. Although extensive research has been conducted on mitigating catastrophic forgetting in general tasks, there is currently a lack of research specifically addressing this issue within the context of complex grounding tasks. Lastly, the original grounding CogVLM only involves in outputting responses with natural languages and BBoxes in [[$x_0$, $y_0$, $x_1$, $y_1$]] format. In this paper, we introduce a new format \{$yaw\_angle$, $pitch\_angle$, $roll\_angle$\} for answering HPE prompts as shown in Figure \ref{fig:abl_rankagg_shot}(b). This enriches the knowledge of the original grounding CogVLM, meanwhile increasing the complexity of output formats. Empirically, we have observed that the directly LoRA~\cite{hu2021lora} fine-tuning and model merging methods frequently generate blended invalid outputs like [[$x_0$, $y_0$, $yaw\_angle$\}, which is referred as invalid answers in this paper. More invalid answers are detailed in Table \ref{tab:head_detection}.  

In this paper, for addressing the catastrophic forgetting problem in grounding tasks, we evaluate and improve the data rehearsal methods~\cite{scialom2022finetuned,huang2024mitigating} that were originally used in non-grounding VLMs. The results show that the visual grounding task, which demands accurate numerical outputs, requires a significantly larger rehearsal ratio than non-grounding VLMs. Here, the rehearsal ratio represents the percentage of images randomly selected from earlier training phases that are reintegrated during the training of new tasks~\cite{scialom2022finetuned,huang2024mitigating}. To improve HPE accuracy and address blended invalid outputs, we propose and validate a model merging method based on LoRA layers. Utilizing this approach, our model demonstrates exceptional robustness, achieving a 31.5\% reduction in Mean Absolute Error (MAE) of Euler angles, compared to the CNN-based state-of-the-art (SOTA) in cross-dataset evaluations. Furthermore, we compare our LoRA layer-based merged model with both the directly LoRA fine-tuned model and the task arithmetic-based merged model within CogVLM. Our approach consistently shows superior performance in both of MAE and invalid answer ratio reduction. Our contributions can be concluded as following:
\begin{itemize}
    \item Our work pioneers the improvement of HPE tasks by directly enhancing the visual grounding capabilities of CogVLM. Rather than utilizing VLMs solely as feature extractors and relying on additional components for pose estimation, we focus on advancing the VLM itself. By directly enabling the VLM to generate pose estimation outputs, we integrate task-specific capabilities within the model, resulting in a streamlined and efficient end-to-end solution.
    \item To the best of our knowledge, this is the first work to explore the issues of catastrophic forgetting and blended invalid response when multiple grounding tasks are involved.
    \item We propose a novel LoRA layer-based model merging method that adopts a ``winner takes all'' strategy, significantly outperforming the CNN-based SOTA, directly LoRA fine-tuned VLM, and task arithmetic-based merged VLM in terms of MAE and invalid answer ratio reduction. This demonstrates our method is able to achieve outstanding robustness and effectiveness in the HPE task, and offers insight into how structured numerical outputs can be integrated into grounding-oriented VLMs.

\end{itemize}

\section{Related Work}

\subsection{Head Pose Estimation (HPE)}

Traditional approaches for HPE include both landmark-based \cite{DBLP:journals/jmlr/King09,DBLP:journals/corr/abs-2403-18104,6939713} and landmark-free methods \cite{DBLP:conf/cvpr/RuizCR18,10973075,DBLP:conf/bmvc/ZhouG20}. Landmark-based methods rely on detecting specific facial landmarks, such as the eyes, nose, and mouth, to estimate head orientation. For example, the Lie Group Kernel method~\cite{6939713} models facial appearance using a linear dynamic representation and compares them in a Lie group space. While such methods can perform well under controlled conditions, they struggles with full-range HPE as extreme poses often obscure facial landmarks. Given our focus on full yaw range HPE, we prioritize landmark-free approaches. Under this approach, several models divide continuous rotation variables into discrete bins for classification purposes~\cite{DBLP:journals/corr/abs-1710-00925,zhou2020whenet,8444061,huang2020improving,zhang2020fdn}. Besides those, FSA-Net~\cite{yang2019fsa} employs a stage-wise regression and feature aggregation scheme to predict Euler angles. Meanwhile, models such as 6DRepNet~\cite{Hempel_2022} and TriNet~\cite{cao2020vectorbased} take a different approach by estimating the rotation matrix instead of directly predicting Euler angles. In this paper, we use several traditional CNN-based landmark-free HPE approaches as baselines. This allows us to assess conventional limitations and benchmark the effectiveness of our proposed method. Beyond CNN-based approaches for HPE, transformer-based designs have also been explored for face-related analysis. For instance, SwinFace~\cite{10216308} leverages a Swin Transformer backbone for 
multi-task face recognition and attribute prediction. However, its focus lies in 
categorical face attributes (e.g., identity, gender, age) rather than continuous 
head pose regression.

\subsection{Grounding in Vision Language Models}

Some VLMs with grounding capabilities can provide accurate BBox information in the format of [[$x_0$, $y_0$, $x_1$, $y_1$]] based on specific prompts~\cite{wang2024cogvlm, bai2023qwenvl,chen2023shikra, you2023ferret}. This functionality is crucial for tasks requiring precise spatial awareness, like object detection and image captioning, as accurately identifying object positions enhances visual scene understanding. For example, Zhai et al.~\cite{10740342} leverage the object detection capabilities of grounding DINO~\cite{liu2023grounding} to improve object counting and enable accurate quantification.  Unlike traditional BBox prediction task, HPE task requires an understanding of 3D spatial relationships to produce accurate Euler angles in the format of \{$yaw\_angle$, $pitch\_angle$, $roll\_angle$\}. Most existing VLMs are not inherently designed to handle such queries, as they typically output 2D BBoxes rather than 3D rotational data.

Nevertheless, some recent efforts have attempted to extend VLMs toward pose and gaze estimation by leveraging their embeddings. For example, CLIP-Gaze~\cite{yin2024clip} applies CLIP to gaze estimation by using frozen encoders with personalized text prompts to generate gaze-irrelevant language features, and then employs a separation loss to push gaze-relevant visual features away from these distractors. Similarly, CLIPose~\cite{10522755} extends CLIP to 6D object pose estimation by combining three separately encoded modalities including, point clouds, RGB images, and text, into a shared space through multi-modal contrastive learning. While effective, both approaches treat CLIP primarily as a fixed feature extractor with auxiliary components, following a “separate encoding + post-hoc alignment” paradigm.
As a result, they rely on cropped inputs, discarding full-scene context. In contrast, our HPE-CogVLM fine-tunes and merges CogVLM so that HPE becomes a native capability of the model. By jointly training prompts and the visual encoder, the model interprets task queries as grounding instructions, locating and estimating head pose directly within full uncropped images. This integrated design uses cross-attention to combine localization and pose estimation in one step, preserving CogVLM’s grounding and language abilities while extending to HPE. Leveraging full-scene RGB inputs, the model remains robust under clutter, multi-person interactions, and extreme or back-facing orientations—key requirements in applications such as driver monitoring, retail analytics, and human–robot interaction. Ultimately, this work moves beyond frozen embeddings toward a unified multimodal solution.

\subsection{Model Merging in LLMs}

There has been extensive exploration of model merging techniques designed to enhance the capabilities of LLMs. These methods aim to combine multiple LLMs, each with specialized functionalities, into a unified model capable of addressing a range of tasks across various domains. The typical merging methods~\cite{ilharco2023editing,wortsman2022model,jang2024model,davari2023model,yadav2024ties,yu2023language,akiba2024evolutionary} usually apply rules or algorithms to trim or merge the parameters of LLMs. For example, task arithmetic~\cite{ilharco2023editing} defines arithmetic rules to incorporate new capabilities or delete undesired ones, allowing task-specific parameters to be adjusted in a controlled manner. More advanced variants introduce sophisticated strategies beyond simple averaging. TIES-merging~\cite{yadav2024ties} addresses parameter conflicts by identifying and pruning contradictory updates, then merging only sign-consistent parameters. Fisher-weighted averaging~\cite{wortsman2022model} leverages Fisher information to weight parameters based on their importance to task performance. Evolutionary merging~\cite{akiba2024evolutionary} employs evolutionary algorithms to search optimal combination strategies across parameter subspaces. Recent work like MoW-Merging~\cite{10900479} introduces dynamic merging with sample-wise adaptive coefficients through gating networks, rather than static merging strategies. While these methods show effectiveness in standard NLP multi-task settings, they remain fundamentally interpolation-based approaches that blend parameters at fine-grained levels. In contrast, our winner-takes-all strategy selects entire LoRA layers based on cosine similarity, avoiding parameter blending altogether. This design prevents invalid mixtures when preserving multiple structured outputs (BBoxes and Euler angles), making it better suited for multimodal grounding tasks.

\subsection{Catastrophic Forgetting Problem}
Catastrophic forgetting has been a significant issue that limits the effectiveness of LLMs, as they tend to forget previously knowledge when learning new knowledge. This problem is particularly pronounced in continual learning settings, where models are sequentially exposed to new tasks and risk losing their ability to perform earlier ones \cite{kirkpatrick2017overcoming}. Various approaches~\cite{10045664,10644076,Kirkpatrick_2017,li-etal-2022-overcoming,xu2018reinforced,huang2021continual,scialom2022finetuned, huang2024mitigating,luo2024empirical,zhang-etal-2023-citb, mok-etal-2023-large} have been developed to address this issue. Ma et al.~\cite{10045664} address catastrophic forgetting by introducing subspace prompt tuning and a novel feature learner to regularize prompt tuning, ensuring that learned prompts retain generalizable features and align with the model's original zero-shot capabilities. Zhang et al.~\cite{10644076} mitigate catastrophic forgetting in continual learning by freezing the text encoder of a VLM and using lightweight task-specific visual adapters to preserve old knowledge while learning new classes without requiring old data. However, those approaches introduce extra parameters, increasing the model’s complexity and memory requirements and their performance remains uncertain for learning multiple grounding tasks with numerical outputs. The rehearsal method~\cite{scialom2022finetuned, huang2024mitigating,luo2024empirical,zhang-etal-2023-citb, mok-etal-2023-large} is the most widely used method to mitigate catastrophic forgetting. It involves reusing a small portion of old task datasets into the new task fine-tuning process. By periodically revisiting earlier knowledge, the model maintains a balanced representation of all tasks, reducing the likelihood of forgetting. Despite these advancements, catastrophic forgetting remains under-explored in grounding tasks, which require precise numerical outputs. In our experiments, we find that grounding tasks require a larger rehearsal ratio than non-grounding VLM tasks to effectively mitigate forgetting. As further discussed in Section~\ref{oprehearsal}, empirical results confirm that ratios of 10\% or higher are necessary to retain grounding ability, whereas non-grounding VLMs often adopt only 1\%~\cite{scialom2022finetuned,huang2024mitigating}. A probable reason is that grounding tasks involve structured numerical outputs (e.g., bounding box coordinates or Euler angles), where even small deviations from the learned distribution can result in invalid predictions. This fragility may explain their stronger reliance on larger rehearsal sets.

\section{HPE-CogVLM Framework} \label{methodology}

\begin{figure*}[t!]
    \centering
    \includegraphics[width=\textwidth, trim=0 80 0 80, clip]{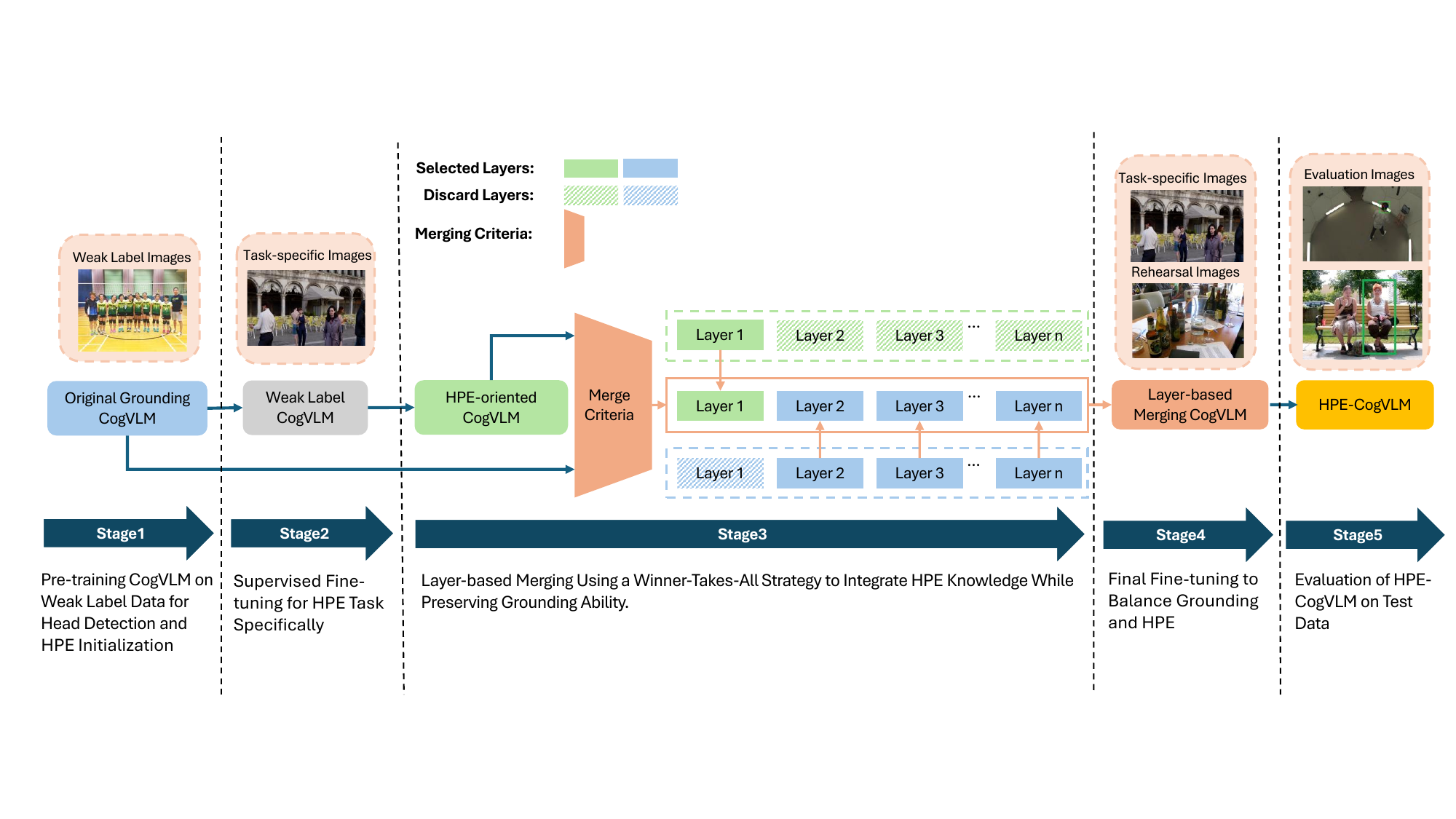}
    \caption{Overview of the multi-stage process for integrating the HPE task into the grounding CogVLM.}

    \label{fig:merge_flow}
\end{figure*}

The proposed framework of HPE-CogVLM as shown in Figure \ref{fig:merge_flow} is structured through a multi-stage process. Each stage of this framework is designed to enhance different aspects of the model's capabilities, gradually refining the parameters to balance HPE and BBox tasks. The fine-tuning process at each stage follows the CogVLM's fine-tuning scripts\footnote{\url{https://github.com/THUDM/CogVLM}}, which implement LoRA~\cite{hu2021lora} across transformer blocks, including the query, key, value of attention layers and dense layers. Subsequently, the LoRA matrices of each layer are accumulated into the corresponding layer in original model. To support this multi-stage process, the framework utilizes a variety of datasets, each serving a distinct role in model training and evaluation. Table \ref{datasets} provides an overview of how these datasets contribute at different stages, such as enhancing human head BBox detection, improving HPE task accuracy, and preventing catastrophic forgetting. Below is a detailed description of each stage in the framework:

\begin{table*}[t!]
  \caption{A detailed overview of various datasets used in our framework.}
  \label{datasets}
  \centering
  \resizebox{\linewidth}{!}{
  \begin{tabular}{l|l|cc|cc|l|l}
    \midrule
    \multirow{2}{*}{\textbf{Task}}   &   \multirow{2}{*}{\textbf{Dataset}}    &   \multicolumn{2}{c}{\textbf{\# of Images}} & \multicolumn{2}{|c|}{\textbf{\# of Heads}} &   \multirow{2}{*}{\textbf{Usage}}  & \multirow{2}{*}{\textbf{Functionality}} \\
    \cmidrule(lr){3-4} \cmidrule(lr){5-6} 
        &   &   \textbf{Train} & \textbf{Test}    & \textbf{Train} & \textbf{Test}  &     \\
    \midrule
    Weak Label Images   &   CrowdHuman~\cite{shao2018crowdhuman}  &   11,731   & - &   94,795 &  -  &   Stage 1  & \begin{tabular}[c]{@{}l@{}} Focus on human head BBox detection\\ Warm-up for HPE task using weak label images\end{tabular}  \\ 
    \midrule
    Task-specific Images    &   Agora~\cite{patel2021agora}   &   9,654    & - & 64,187  &   - &   Stage 2, 4 & \begin{tabular}[c]{@{}l@{}}Training with precise HPE labels from synthetic images\\ Focus on HPE task to improve accuracy\end{tabular} \\
    \midrule
    \multirow{3}{*}{Rehearsal Images}   &   Refcoco~\cite{yu2016modeling} &   42,404   &   -   & -   & -  &   \multirow{3}{*}{Stage 4}   &\multirow{2}{*}{\raggedright \begin{tabular}[c]{@{}l@{}}Preventing catastrophic forgetting with Refcoco/+/g training set\\ \\ Investigating rehearsal ratio optimization\end{tabular}}\\
    &   Refcoco+~\cite{yu2016modeling} &   42,278   &   -   &  -  & -  &    \\
    &   Refcocog~\cite{mao2016generation} &   42,226   &   -   &  -  & -  &     \\
    \midrule
    \multirow{4}{*}{Evaluation Images} &   CMU Panoptic~\cite{joo2016panoptic}    &   -  & 16,216   & -  &   32,738   &  \multirow{4}{*}{Stage 5} &\multirow{4}{*}{\raggedright \begin{tabular}[c]{@{}l@{}}Evaluating the accuracy of the HPE task on CMU Panoptic test set\\ \\ Evaluating the BBox detection knowledge on Refcoco/+/g test set \end{tabular}} \\
    &   Refcoco~\cite{yu2016modeling} &   -   &   3785   & -   & -  &      \\
    &   Refcoco+~\cite{yu2016modeling} &  -   &   3773   &  -  & -  &     \\
    &   Refcocog~\cite{mao2016generation} &  -   &   5023   &  -  & -  &     \\
    \bottomrule
  \end{tabular}
  }
\end{table*}

\subsection{Stage 1: Pre-training of the Original Grounding CogVLM on Weak Label Data} 
\label{pre-train}

The CrowdHuman dataset \cite{shao2018crowdhuman} contains a large number of real human head instances across diverse poses and contexts, offering strong supervision for BBox learning. Since CrowdHuman lacks ground-truth pose annotations, we infer weak HPE labels using 6DRepNet. The resulting weak label CogVLM thus combines abundant real-image head data with approximate pose cues, establishing a solid foundation for subsequent HPE tasks.

\subsection{Stage 2: Supervised Fine-tuning of the Weak Label CogVLM on Task-specific (HPE) Data} \label{fine-tune}

Following pre-training, the model undergoes supervised fine-tuning on task-specific HPE data using the Agora dataset. Unlike CrowdHuman's real images with weak labels, Agora provides synthetic images with precise head pose annotations, enabling the model to learn fine-grained orientation predictions. The output model is referred as the HPE-oriented CogVLM as shown in Figure \ref{fig:merge_flow}. This refined model combines the real-image exposure and contextual background knowledge gained from the previous stage with the task-specific precision acquired in this stage.

\subsection{Stage 3: Layer-based Merging between Original Grounding CogVLM and HPE-oriented CogVLM}
 
During this key stage, the original grounding CogVLM is merged with the HPE-oriented CogVLM (from Stage 2) using a cosine similarity criterion. After
LoRA fine-tuning, all LoRA updates are merged back into the base weights, so
each projection layer (e.g., the $W_q$, $W_k$, $W_v$ matrices in attention
and the up/down projections in the MLP) is represented by a single consolidated
weight matrix $W \in \mathbb{R}^{d \times k}$. Cosine similarity between
corresponding projection layers of the two models is computed directly on these
matrices. Concretely, we use \texttt{F.cosine\_similarity} with
\texttt{dim = -1}, which computes cosine similarity along the last dimension
(the $k$-dimensional feature axis) and produces a length-$d$ vector of
row-wise similarities. The final similarity score for the projection layer is
obtained by averaging this vector. Bias vectors are not included in this process, as LoRA adapts only weight matrices and they dominate each layer's representation. Cosine similarity is used to gauge the amount of information shared between layers. The threshold of cosine similarity can help to determine whether layers from the HPE-oriented CogVLM should be integrated into the final model. Since LoRA finetuning is applied in previous stages, most of the original
model parameters are only minimally altered, necessitating a high threshold for cosine similarity. In our experiments, We empirically set the threshold at 0.95. If the similarity falls below this threshold, we opt to completely retain the
original knowledge. Otherwise, if the similarity exceeds the threshold, which indicates a substantial overlap in information due to the stringent criteria, we select the entire layer from the HPE-oriented CogVLM to guarantee the minimal risk
of losing important existing knowledge. Consequently, the merging criteria is detailed as below:

\begin{itemize}

\item \textbf{Safeguard (lowest 1\%).} 
We calculate and rank the cosine similarities across all layers from both models, and always select the layer from the original grounding CogVLM model within the smallest 1\% of cosine similarities.

\item \textbf{Main rule — threshold-based selection.} When the cosine similarity between two layers from each model is less than the threshold, we also select the layer from the original grounding CogVLM.

\item \textbf{Otherwise}, we choose the layer from the HPE-oriented CogVLM.

\end{itemize}

Precedent methods usually apply either direct LoRA fine-tuning or parameter-level merging strategies that rely on hyperparameter heuristics to discard and combine weights~\cite{ilharco2023editing, yadav2024ties,yu2023language,akiba2024evolutionary}. In practice, we find that direct LoRA fine-tuning alone cannot reach the desired HPE accuracy, while interpolation-based merging can improve accuracy but fails under multiple grounding tasks with heterogeneous output formats. A common failure case is blended outputs: when queried with HPE prompts, the merged model sometimes returns irrelevant NLP text (e.g., “a person of head”) or malformed structures (e.g., [[999,231,123,389\}). More examples are provided in Table~\ref{tab:head_detection}.

To overcome these issues, we adopt a “winner-takes-all” policy that selects entire layers from either the original grounding CogVLM or the HPE-oriented CogVLM as shown in Figure \ref{fig:merge_flow}. This design ensures that only layers with substantial informational overlap are replaced, preserving the original model’s grounding capability while selectively integrating HPE knowledge. As a result, existing knowledge remains intact even when layers are borrowed from the HPE-oriented model.

This approach offers two key advantages: (1) It helps mitigate attention subspace interference and reduces the risk of invalid blended outputs by preserving the structural integrity of each layer. In grounding-oriented VLMs, attention projection matrices and MLP projections define coherent feature subspaces that are closely tied to structured output generation. When parameters from two task-specialized models are partially interpolated, their learned subspaces may become entangled. This risk is especially pronounced when the tasks involve heterogeneous structured outputs, such as BBox vs. Euler angles. The entangled subspaces may cause the model to simultaneously activate competing output templates. Then this competition can destabilize token-level decoding, ultimately increasing the likelihood of mixed-format responses. Therefore, whole-layer selection preserves the internal representation structure of each layer. (2) It provides a more stable initialization for downstream rehearsal fine-tuning. Since whole-layer selection reduces representational conflicts at initialization, the merged model requires less fine-tuning to restore task balance. In our experiments, Stage 4 fine-tuning converges effectively within a short training budget (less than one epoch), suggesting that the merged initialization is already well-structured for subsequent adaptation.

\subsection{Stage 4: Continual Fine-tuning of Layer-based Merging CogVLM on Mixture Data}\label{2ndfine-tune}

After merging, the model undergoes brief continual fine-tuning (less than one epoch) on mixed data: task-specific HPE images plus rehearsal images. The optimal rehearsal ratio is pre-determined in Stage 1 by testing proportions of 0\%, 1\%, 10\%, and 25\%. The rationale for incorporating additional brief
fine-tuning is that while layer merging preserves parameter integrity, it lacks the task-specific refinement needed for accurate HPE. This short fine-tuning restores grounding balance and boosts prediction accuracy, yielding the final HPE-CogVLM (Figure \ref{fig:merge_flow}).

\subsection{Stage 5: Evaluation of HPE-CogVLM on Test Data}

To demonstrate the robustness of our model, we utilize real-world CMU Panoptic images to evaluate the model's performance on the HPE task. Meanwhile, we use rehearsal test datasets to assess the model's performance on the BBox prediction task.

\vspace{11pt}

Compared to directly LoRA fine-tuning and some model merging methods, this framework ensures that HPE-CogVLM maintains a high level of accuracy in HPE task, effectively mitigating the issues of blended invalid outputs and catastrophic forgetting problem. Experimental results will be shown in the section \ref{res}.

\section{Experiments Setup} 
\label{exp setup}
\subsection{HPE Task Prompt Design}

\begin{table*}[t!]
\caption{Prompts and Responses Design for HPE Task.}
\centering
    \begin{tabular}{ll}
    \toprule
    \multicolumn{2}{c}{\textbf{BBox Prediction Task}} \\
    \midrule
    \textbf{Designed Prompt} &   How many human heads are in this image and what are the head bounding boxes?    \\
    \midrule
    \textbf{Correct Answer} &   Their head bounding boxes are [[106,168,148,242;245,168,270,230]].    \\
    \midrule
    \multirow{5}{*}{\textbf{Invalid Answer (Reason)}} &    [[000,111,222,333...    (\textit{Recycled output error}) \\
        &   \{112,432,211\}   (\textit{Angle format output error})   \\
        &   A man in Red    (\textit{NLP output error})    \\
        &   [[212,123,212\}  (\textit{Mixed output error})  \\
        &   [[234,134,100,111]] (\textit{Logical error})   \\
    \midrule
    \multicolumn{2}{c}{\textbf{Head Pose Estimation Task}} \\
    \midrule
    \textbf{Designed Prompt} &  What is the head yaw pitch roll inside the bounding box [[106,168,148,242]]?  \\
    \midrule
    \textbf{Correct Answer} &   The head orientation angles are \{072,354,002\}.   \\
    \midrule
    \multirow{5}{*}{\textbf{Invalid Answer (Reason)}} &   \{112,432,211,201\}   (\textit{Wrong number error})   \\
        &   [[234,134,100,111]] (\textit{BBox format output error})  \\
        &   A person head (\textit{NLP output error})  \\
        &   [[212,123,212\} (\textit{Mixed output error})  \\
        &   \{999,389,001\} (\textit{Logical error})  \\
    \bottomrule
    \end{tabular}
\label{tab:head_detection}
\end{table*}

In some traditional CNN-based models, such as 6DRepNet and HopeNet, cropping the human head region is required as the initial step. In this paper, a new prompt method is proposed, allowing us to train HPE task utilizing the information of full images. In our prompts, BBox coordinates are leveraged to specify the human head of interest when multiple people are present. Therefore, the system is capable of effectively focusing on specific heads, which makes it easier to reduce the need for labor-intensive manual annotations and automate the inference process. Meanwhile, the global features from self-attention and head of interest features from cross-attention are both learnt to improve the robustness of HPE task. Figure \ref{fig:abl_rankagg_shot} (b) illustrates a sample of our custom-designed prompts and responses tailored for the HPE task, demonstrating how the system interprets and responds to specific queries. Additionally, table \ref{tab:head_detection} provides more examples of prompts and responses for the HPE and BBox prediction tasks. The BBox format adheres to the specifications set by CogVLM~\cite{wang2024cogvlm}. 

For head pose, we employ Euler angles (yaw, pitch, roll) as the output representation. Our VLM predicts each angle as discrete tokens trained with cross-entropy loss, which naturally avoids the numerical discontinuities of regression. Alternative representations such as quaternions or rotation matrices introduce challenges for token-based generation. Quaternions require a unit-norm constraint, and rotation matrices impose a strict orthogonality constraint, both of which are difficult to enforce strictly during discrete token generation. While 6D rotation representations~\cite{Hempel_2022} were specifically proposed to eliminate these hard geometric constraints, they increase output dimensionality, demand higher numerical precision, and reduce interpretability compared to Euler angles.

Euler angle labels are first converted to positive floats. These values are then rounded to the nearest integer, formatted as strings with a fixed length of three characters, padded with zeros where necessary. This table also includes several typical examples of invalid answers to highlight how invalid answers can lead to completely ineffective outputs when multiple grounding tasks requires accurate numerical output varied in range and quantity. In response to these issues, we define a new metric described in Section \ref{metric} to assess the model's availability.

\subsection{Datasets}

Table \ref{datasets} outlines datasets used in various stages of our framework. The CrowdHuman dataset~\cite{shao2018crowdhuman} serves as the pre-training dataset due to its extensive collection of human images. Its head pose annotations are derived from pseudo-labels inferred by the pre-trained 6DRepNet~\cite{Hempel_2022} model, and thus are referred as weak label images. It enables the model to obtain the capability of detecting real human heads and warm up the HPE task in stage 1. The synthetic Agora dataset~\cite{patel2021agora} serves as the task-specific HPE images, which encompasses full-range of human head yaw angle images and provides the ground-truth of SMPL-X parameters~\cite{pavlakos2019expressive}. Its head pose annotations are generated using the method of DirectMHP~\cite{zhou2023directmhp}\footnote{\url{https://github.com/hnuzhy/DirectMHP}}. Following the DirectMHP pipeline, pose estimation is normalized with camera parameters, and both annotations and predictions are aligned to a shared canonical head reference defined by a common virtual head model. This ensures consistent coordinate conventions across AGORA and CMU Panoptic and enables fair cross-dataset evaluation. This dataset is used in stages 2 and 4 of our framework to enhance HPE accuracy. The Refcoco~\cite{yu2016modeling}, Refcoco+~\cite{yu2016modeling}, and Refcocog~\cite{mao2016generation} train datasets, which are originally utilized by CogVLM to learn BBox prediction, are chosen as rehearsal images to help mitigate the catastrophic forgetting of existing BBox capability. In our experiments, various portions of the rehearsal images are applied to determine the optimal rehearsal ratio~\cite{scialom2022finetuned,huang2024mitigating}. 

For evaluation, a subset of the CMU Panoptic dataset serves as the test dataset for evaluating HPE task, as its panoptic images of real people closely mirror real-life scenarios. The selection of images and labels follows the DirectMHP approach~\cite{zhou2023directmhp}. To evaluate object BBox localization, the test datasets including testA and testB data from Refcoco and Refcoco+, as well as the test dataset from Refcocog, are selected as the BBox evaluation datasets.

\subsection{Implementation Details} \label{implemen}

Throughout the LoRA fine-tuning process, a LoRA rank of 10 is used. The learning rate of \num{1e-4} is used in pre-training stage. All other training parameters follow the default settings of the CogVLM. The experiments are performed on using two NVIDIA A100 80GB GPUs with a training batch size of 8. The training processes in stages 1, 2, and 4 of our framework cost 20, 50, and 10 hours, respectively.

\subsection{Evaluation Metrics} \label{metric}

We define four evaluation metrics for assessing HPE and BBox prediction tasks as follows:

\textbf{Angle Error Ratio} ($\text{E}_\text{angle}$): $\text{E}_\text{angle}$ = \(\frac{e_\text{angle}}{t_\text{angle}}\), where $e_\text{angle}$ denotes the number of invalid HPE answers and $t_\text{angle}$ denotes the number of total HPE answers. This new metric is defined to assess the capability of models to provide relevant numerical outputs for HPE task. When we prompt with a HPE query, the CogVLM could produce irrelevant responses such as an natural language processing (NLP) task response like ``a person head'', a BBox task response like ``[[111,222,333,444]]'', or a blended response like ``[[111,999,999,99\}'' as shown in Table \ref{tab:head_detection}. 

\textbf{BBox Error Ratio} ($\text{E}_\text{bbox}$): $\text{E}_\text{bbox}$ = \(\frac{e_\text{bbox}}{t_\text{bbox}}\), where $e_\text{bbox}$ denotes the number of invalid BBox answers and $t_\text{bbox}$ denotes the number of total BBox answers. This new metric is defined to assess the capability of models to provide relevant numerical outputs for BBox prediction task.

\textbf{BBox accuracy} (ACC.): Acc. = \(\frac{m}{\hat{m}}\), where $m$ denotes the number of valid BBox answers with \(\text{IoU} > 0.5\) and $\hat{m}$ denotes the number of total valid BBox answers. A BBox prediction is considered to be accurate if the intersection over union (IoU) between the ground-truth and the prediction exceeds 0.5~\cite{yu2016modeling}. And the invalid answers are excluded from accuracy and MAE calculation.
    
\textbf{MAE of Euler angles} (MAE): For the HPE task, the MAE between the ground-truth Euler angles and the predicted Euler angles is defined as follows:

\begin{equation}
    \text{MAE} = \frac{1}{n} \sum_{i=1}^{n} \min(360^\circ - |\hat{A}_i - A_i|,|\hat{A}_i - A_i|)
\end{equation}
where $\hat{A}_i$ represents the ground-truth's Euler angles, $A_i$ represents the predicted Euler angles, and variable $n$ denotes the number of valid HPE answers. The MAE is measured in a circular manner rather than linearly, leading to the inclusion of a term that minimizes the difference between the predicted and actual angle by considering a full 360-degree rotation~\cite{Hempel_2022, zhou2023directmhp}. For example, a prediction of $359^\circ$ against a ground-truth of $1^\circ$ should yield an error of $2^\circ$, not $358^\circ$ as in linear MAE. Therefore, the circular MAE avoids overestimating errors at the wrap-around boundary. In this paper, the MAE value is considered as the average of MAE for yaw, pitch and roll Euler angles.

\subsection{Baseline Methods} \label{baselinedescription}

In this paper, several types of baseline methods are considered to be compared with our HPE-CogVLM.

\textbf{Traditional CNN-based models}, including 6DRepNet, HopeNet and WHENet, serve as the baselines for CNN-based approaches. The 6DRepNet model, recognized as the current SOTA, is specifically retrained and tested on the same Agora and CMU datasets used in the VLM experiment to ensure a fair comparison. This model is trained 100 epochs, and the best MAE is selected for baseline analysis with our HPE-CogVLM. The pre-trained models of HopeNet and WHENet are utilized because HopeNet scripts are hard-coded and WHENet training scripts are not publicly available.

\textbf{PoseBERT}, a transformer-based module for pose sequence modeling~\cite{leveraging_mocap}, is included as a baseline. It applies self-attention to refine pose sequences and operates as a plug-and-play temporal module on top of image-based estimators, in contrast to our integrated VLM approach. PoseBERT predicts SMPL parameters, whereas our pipeline uses SMPL-X; therefore, we follow the official SMPL-X transfer utilities\footnote{\url{https://github.com/vchoutas/smplx/tree/main/transfer_model\#smpl-to-smpl-x}} to map SMPL outputs into the SMPL-X joint convention before deriving yaw, pitch, and roll from the head-joint orientation. In addition, PoseBERT is restricted to refining the pose of a single detected person. We employ the publicly released pretrained model, as training scripts are not available.

\textbf{Frozen CogVLM+Regressor}, where the entire CogVLM (vision and language encoders) is frozen, and a lightweight two-layer MLP regressor (\emph{feature embedding} $\rightarrow$ 512 $\rightarrow$ 3 for yaw/pitch/roll) is attached on top of the hidden representations from the final multimodal transformer block after cross-attention. Two variants are considered to examine the effect of feature aggregation:
(A) the regressor takes the mean-pooled \emph{visual tokens} representations only;
(B) the regressor takes the mean-pooled representations of \emph{both visual and text tokens}. The model takes full-scene images with grounding-style prompts, identical to our HPE-CogVLM. The regressor is trained for 100 epochs using the WHENet wrapped loss~\cite{zhou2020whenet}, and the best test MAE is reported.

\textbf{Vision Tower(ViT)+Regressor}, where only the CogVLM vision tower is used as a frozen feature extractor. Since CogVLM removes the final ViT layer that aggregates the \([CLS]\) representation for contrastive learning, no \([CLS]\) token is available~\cite{wang2024cogvlm}. Therefore, we apply the same two-layer MLP regressor to the mean-pooled visual features from the vision tower.  To facilitate convergence, cropped head images are used as inputs. This baseline reflects the common practice of leveraging frozen visual embeddings with a lightweight predictor, without multimodal fine-tuning. The regressor is trained for 100 epochs using the WHENet wrapped loss, and the best test MAE is reported.

\textbf{Non-merging CogVLM}, directly LoRA fine-tuned model without applying model merging technique, serves as a comparison point to evaluate the effectiveness of our merging approach versus the LoRA fine-tuning only method~\cite{scialom2022finetuned,huang2024mitigating, luo2024empirical}. The difference of Non-merging CogVLM and our HPE-CogVLM methods is that the Non-merging CogVLM bypasses stages 2 and 3, instead it undergoes significantly more training iterations in stage 4 which is equal to the total iterations of stages 2 and 4 in the HPE-CogVLM framework. For examples, our HPE-CogVLM is fine-tuned 25k and 5k iterations in stages 2 and 4 respectively, while the Non-merging CogVLM is solely fine-tuned in stage 4 for 30k iterations. This ensures fair comparison with respect to HPE task training iterations. 

\textbf{Task Arithmetic (TA) merging CogVLM}, which adheres to our framework but replacing the layer-based merging with TA based merging, is to provide a baseline for comparing our merging approach with another merging method. The TA merging process is chosen as it forms the foundation for many other merging algorithms~\cite{yadav2024ties,yu2023language}. In this process, we set the lambda parameter of task arithmetic to 0.5~\cite{ilharco2023editing}, assigning equal importance to both the BBox prediction task and the HPE task.\\

\section{Experimental Results}\label{res}

\subsection{Baseline Comparison} \label{maincompare}

\begin{table*}[t!]
\caption{Comparison of HPE-CogVLM performance with various baselines. The best results are highlighted in bold.}
\centering
\begin{tabular}{l|cc|cc|cc|cc}
\toprule
\multirow{2}{*}{\textbf{Model}} & \multicolumn{2}{c}{\textbf{Refcoco}} & \multicolumn{2}{c}{\textbf{Refcoco+}} & \multicolumn{2}{c}{\textbf{Refcocog}} & \multicolumn{2}{c}{\textbf{CMU Panoptic}} \\
\cmidrule(lr){2-3} \cmidrule(lr){4-5} \cmidrule(lr){6-7} \cmidrule(lr){8-9} 
 & \textbf{$\text{Acc}_\text{test}$↑} & \textbf{$\text{E}_\text{bbox}$↓} & \textbf{$\text{Acc}_\text{test}$↑} & \textbf{$\text{E}_\text{bbox}$↓} & \textbf{$\text{Acc}_\text{test}$↑} & \textbf{$\text{E}_\text{bbox}$↓} & \textbf{$\text{MAE}_\text{test}$↓} & \textbf{$\text{E}_\text{angle}$↓} \\
\midrule
WHENet & - & - & - & - & - & - & 29.55 & - \\
HopeNet & - & - & - & - & - & - & 22.16 & - \\
6DRepNet & - & - & - & - & - & - & 10.74 & - \\
PoseBERT & - & - & - & - & - & - & 22.32 & - \\
Frozen CogVLM + Regressor (visual tokens) & - & - & - & - & - & - & 32.11 & - \\
Frozen CogVLM + Regressor (all tokens) & - & - & - & - & - & - & 33.47 & - \\
Vision Tower (ViT) + Regressor & - & - & - & - & - & - & 10.88 & - \\
Original Grounding CogVLM & 91.4\% & 0\% & 86.7\% & 0\% & 90.2\% & 0\% & - & - \\
Non-merging CogVLM & 91.1\% & 0\% & 85.2\% & 0\% & 88.9\% & 0\% & 8.18 & 0.13\% \\
TA merging CogVLM & 89.5\% & 0\% & 82.3\% & 0\% & 86.1\% & 0\% & 7.72 & 68.9\% \\
HPE-CogVLM   &   90.5\% & 0\% & 84.7\% & 0\% & 87.8\% & 0\% & \textbf{7.36} & \textbf{0.052\%} \\

\bottomrule
\end{tabular}
\label{tab:model_performance}
\begin{flushleft}
\scriptsize Note: A dash (``-'') indicates that the model is either not capable of performing the specified task or not applicable to the specified metric. For instance, WHENet, HopeNet, and 6DRepNet only accept detected human head bounding boxes for HPE prediction, meaning they cannot perform bounding box detection. As a result, their performance on the Refcoco, Refcoco+, and Refcocog tasks is marked with a ``-''. This applies to all the tables in the following results subsections.
\end{flushleft}

\end{table*}

The results in Table \ref{tab:model_performance} show the performance comparison between our HPE-CogVLM and various baselines described in Section \ref{baselinedescription}. In comparison with traditional CNN-based models, our HPE-CogVLM presents the significantly lower MAE. The HPE-CogVLM MAE is 75.1\%, 66.8\%, and 31.5\% lower than WHENet, HopeNet, and 6DRepNet, respectively. 

For the transformer-based baseline, the performance gap is also clear: PoseBERT achieves 22.32 MAE on CMU Panoptic, far underperforming our HPE-CogVLM at 7.36. This corresponds to a 67\% reduction in error, indicating that despite its transformer backbone, PoseBERT is less effective than our integrated VLM approach. 

For regression head baselines using CogVLM features, both variants of Frozen CogVLM+Regressor yield comparable results: visual tokens-only (32.11 MAE) versus combined visual and text tokens (33.47 MAE). The minimal difference suggests that cross attention has already integrated prompt-relevant information into the visual representations, making the text representations in subsequent hidden states redundant for the pooling operation. By contrast, Vision Tower+Regressor achieves 10.88 MAE by using cropped inputs and pure visual features, and is competitive with strong CNN baselines such as 6DRepNet. Nevertheless, all frozen-feature approaches fall short of our fully fine-tuned HPE-CogVLM (7.36 MAE). These results indicate that although CogVLM’s pre-trained embeddings contain useful visual cues, frozen multimodal features cannot provide the fine-grained spatial grounding required for reliable pose estimation in multi-person scenes. Robust HPE requires tighter integration between localization and orientation estimation, which fully training facilitates more effectively than frozen feature approaches.

In comparison with Non-merging CogVLM, HPE-CogVLM MAE is 10\% lower than the Non-merging CogVLM. Meanwhile the \textbf{$E_\text{angle}$} of our model is 2.5 times smaller than the Non-merging CogVLM. This indicates that our LoRA layer-based merging method is more proficient in HPE than the method that does not utilize any model merging technique. Regarding the BBox results, the HPE-CogVLM's BBox prediction accuracy in test datasets is 0.6\%, 0.5\% and 1.1\% lower than the Non-merging CogVLM, however, the Non-merging CogVLM costs five times more iterations for training rehearsal images as discussed in Section \ref{baselinedescription}. This demonstrates that even with only 1/5 of the rehearsal image training iterations, our HPE-CogVLM achieves a comparable level of BBox accuracy.

Comparing with TA merging CogVLM, our HPE-CogVLM wins in all the metrics. For instance, when evaluated on test datasets, the BBox prediction accuracy of the HPE-CogVLM exceeds that of the TA merging CogVLM by 1\%, 2.4\%, and 1.7\%, respectively. For the HPE task performance, \textbf{$E_\text{angle}$} of TA merging CogVLM is 68.9\%, which is 1325 times larger than that of HPE-CogVLM, indicating that only 31.1\% of the responses for the HPE task are valid. Due to the high number of invalid HPE responses, the MAE metric becomes ineffective for assessing the performance. This highlights that even with an additional round of fine-tuning in stage 4, the TA merging fails to produce relevant numerical responses within our research domain, ultimately proving ineffective for the HPE task. In contrast, by applying our LoRA layer-based merging method, HPE-CogVLM successfully achieves the lowest MAE and invalid output ratio, demonstrating the superiority of this approach.

\subsection{Catastrophic Forgetting Pattern in HPE Task}

\begin{table}[t!]
\caption{The impact of catastrophic forgetting when no data rehearsal are applied.}
\centering
\begin{tabular}{ccccc}
\toprule
\textbf{Iterations} & \textbf{$\text{Acc}_\text{test}$↑} & \textbf{$\text{E}_\text{bbox}$↓} & \textbf{$\text{MAE}_\text{test}$↓} & \textbf{$\text{E}_\text{angle}$↓} \\
\midrule
0 & 91.4\% & 0\% & - & - \\ 
100 & 91.3\% & 0\% & - & - \\ 
500 & 28.1\% & 36.2\% & 41.16 & 2.5\% \\ 
1000 & 10.8\% & 10.2\% & 42.16 & 0.1\% \\ 
\bottomrule
\end{tabular}

\label{tab:research of Cf}
\end{table}

Table \ref{tab:research of Cf} illustrates the profound impact of catastrophic forgetting in a model trained for HPE task only using Agora dataset. The Refcoco test accuracy starts at a high of 91.4\% at iteration 0, indicating initial proficiency in object detection. As the number of training iteration increases and the model is increasingly exposed to the HPE task, the Refcoco test accuracy drastically decreases to 10.8\% at iteration 1000. This sharp decline illustrates significant forgetting of the original BBox knowledge. \textbf{$\text{E}_\text{bbox}$} rises significantly from 0\% to 36.2\% at iteration 500 and then decreases to 10.2\% at iteration 1000. This trend suggests that the model initially adapts to the new task at the expense of previously learned behaviors, causing a temporary increase in errors before stabilizing. The MAE improves from "Not capable" at iteration 100 to 42.16 at iteration 1000, indicating that the model begins to gain proficiency in the new task. The decline in \textbf{$E_\text{angle}$} from 2.5\% to 0.1\% implies that the model's HPE output format becomes more consistent over time.

What is particularly noteworthy in this scenario is the nature of forgetting and learning displayed by the model—old knowledge is significantly diminished before new knowledge is solidified. This contrasts with human learning process, where new and old knowledge often coexist and can even enhance each other’s acquisition. In human cognition, learning new tasks frequently involves integrating new information with existing knowledge, without the catastrophic forgetting seen in this model.


\subsection{Selecting Optimal Rehearsal Ratios for Mitigating the Catastrophic Forgetting Problem} \label{oprehearsal}

\begin{table}[t!]
\caption{Performance of weak label CogVLM under various rehearsal ratios.}
\centering
\resizebox{\linewidth}{!}{
\begin{tabular}{cccccc}
\toprule
\textbf{Iterations} & \textbf{Rehearsal Ratio} & \textbf{$\text{Acc}_\text{test}$↑} & \textbf{$\text{E}_\text{bbox}$↓} & \textbf{$\text{MAE}_\text{test}$↓} & \textbf{$\text{E}_\text{angle}$↓} \\
\midrule
0k & 0\% & 91.4\% & 0\% & - & - \\
10k & 0\% & 21.8\% & 0.026\% & 17.20 & 0.48\% \\
10k & 1\% & 77.5\% & 0.19\% & 21.51 & 0.85\% \\
10k & 10\% & 91.0\% & 0\% & 19.32 & 0.32\% \\
10k & 25\% & 91.5\% & 0\% & 19.92 & 0.23\% \\

\bottomrule
\end{tabular}
}
\label{tab:rehearsal}
\end{table}

Table \ref{tab:rehearsal} presents the performance of weak label CogVLM across different proportions (0\%, 1\%, 10\%, 25\%)~\cite{scialom2022finetuned,huang2024mitigating, luo2024empirical} of the rehearsal dataset in stage 1. The primary aim is to determine the appropriate data rehearsal ratio to retain old knowledge for the fine-tuning in stage 4. The Refcoco test accuracy at iteration 0 is 91.4\%, indicating proficiency with the BBox prediction tasks. After the training is finished, the results demonstrate a clear trend that as the rehearsal ratio increases, the Refcoco test accuracy substantially improves. Starting at a low of 21.8\% when no Refcoco data is used, the accuracy spikes to 77.5\% with just 1\% of rehearsal ratio, eventually reaching over 91\% with 10\% and 25\% of rehearsal ratio. This clearly shows that the more original task data used in learning a new task, the less catastrophic forgetting occurs. In the \textbf{$\text{E}_\text{bbox}$} column, the consistently low \textbf{$\text{E}_\text{bbox}$} values suggest that the availability of BBox predictions tend to stabilize after 10K iterations. MAE and \textbf{$E_\text{angle}$} for HPE task show a fluctuating trend. Since the head pose weak label is provided for this pre-training stage, they may not fully reflect the model’s true HPE performance. Rehearsal ratios of 10\% and 25\% are selected for the stage 4 experiment due to high refcoco BBox prediction accuracy. These ratios are significantly higher than the commonly used 1\% rehearsal ratio in non-grounding tasks~\cite{scialom2022finetuned,huang2024mitigating}.

\subsection{The Influence of Rehearsal Ratios on Multiple grounding task Learning} \label{ab2}

\begin{figure}[t!]
\small
\centering
\begin{minipage}{\linewidth}
\centering
\includegraphics[width=0.9\linewidth]{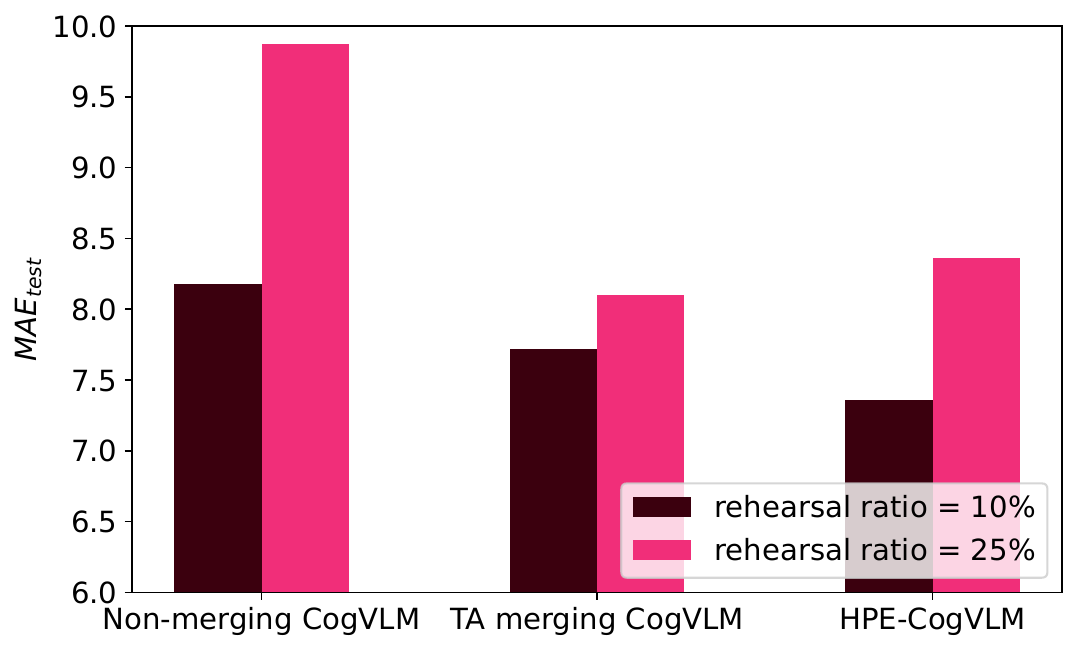}
\centering\mbox{\small (a) The influence of rehearsal ratio on MAE.}
\end{minipage}
\begin{minipage}{\linewidth}
\centering
\includegraphics[width=0.9\linewidth]{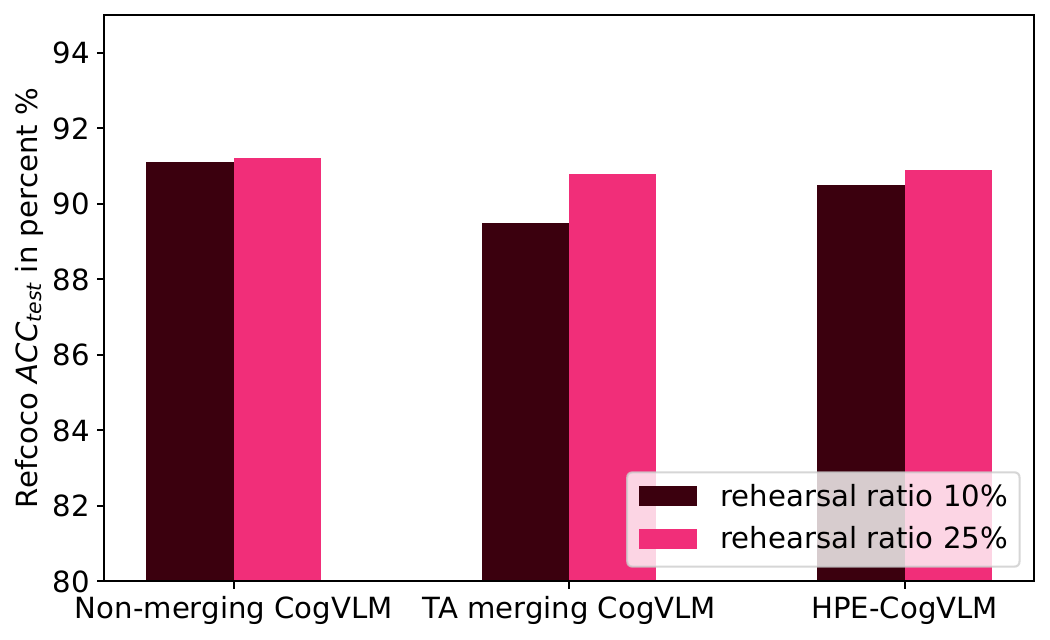}
\centering\mbox{\small (b) The influence of rehearsal ratio on BBox Prediction.}
\end{minipage}
\caption{The model performance under various rehearsal ratios (10\% and 25\%). (a) shows the MAE results under rehearsal ratio 10\% and 25\% on VLMs. (b) shows the Refcoco Test BBox accuracy results under rehearsal ratio 10\% and 25\% on VLMs.
}
\label{fig:influence of rehearsal ratio}
\end{figure}
 
Figure \ref{fig:influence of rehearsal ratio} presents comparative results for both the BBox prediction task and the HPE task under two different rehearsal ratios in stage 4. Between the two HPE-CogVLM models, the one with a lower rehearsal ratio (10\%) achieves a better MAE of 7.36, which is 12\% lower than the 8.36 observed with the higher (25\%) ratio. Conversely, the Refcoco test accuracy improves slightly with higher rehearsal ratios, showing increases of 0.3\% compared to the lower ratio. The similar phenomenon also presents in Non-merging CogVLM and TA merging CogVLM results. The results clearly show that a higher rehearsal ratio helps retain existing knowledge by incorporating more data from previous tasks into fine-tuning, but this comes at the cost of new task performance. So we seek for balance between the retention of old knowledge and the performance on new tasks. In our case, the 10\% rehearsal ratio achieves significantly better HPE performance, while the BBox prediction is only slightly better with the 25\% rehearsal ratio. After balancing both factors, the HPE-CogVLM trained with the 10\% rehearsal ratio is chosen as the optimal model.

\subsection{The Influence of Cosine Similarity Threshold on HPE-CogVLM}

\begin{table*}[t!]
\caption{Sensitivity analysis of HPE-CogVLM performance under different cosine similarity thresholds.}
\centering
\begin{tabular}{p{1.2cm}|cc|cc|cc|cc}
\toprule
\multirow{2}{*}{\textbf{Cos. Sim.}} & \multicolumn{2}{c}{\textbf{Refcoco}} & \multicolumn{2}{c}{\textbf{Refcoco+}} & \multicolumn{2}{c}{\textbf{Refcocog}} & \multicolumn{2}{c}{\textbf{CMU Panoptic}} \\
\cmidrule(lr){2-3} \cmidrule(lr){4-5} \cmidrule(lr){6-7} \cmidrule(lr){8-9} 
 & \textbf{$\text{Acc}_\text{test}$↑} & \textbf{$\text{E}_\text{bbox}$↓} & \textbf{$\text{Acc}_\text{test}$↑} & \textbf{$\text{E}_\text{bbox}$↓} & \textbf{$\text{Acc}_\text{test}$↑} & \textbf{$\text{E}_\text{bbox}$↓} & \textbf{$\text{MAE}_\text{test}$↓} & \textbf{$\text{E}_\text{angle}$↓} \\
\midrule
0.7 & 83.8\% & 0.053\% & 78.2\% & 0.080\% & 81.1\% & 0.060\% & 6.98 & 0.049\% \\
0.8 & 86.4\% & 0\% & 82.4\% & 0\% & 83.9\% & 0\% & 7.12 & 0.049\% \\
0.9 & 90.0\% & 0\% & 84.3\% & 0\% & 85.7\% & 0\% & 7.32 & 0.052\% \\
0.95 &   90.5\% & 0\% & 84.7\% & 0\% & 87.8\% & 0\% & 7.36 & 0.052\% \\
0.98 & 90.7\% & 0\% & 84.7\% & 0\% & 88.0\% & 0\% & 7.83 & 0.061\% \\

\bottomrule
\end{tabular}
\label{tab:csthreshold}
\caption*{\footnotesize Note: Cos. Sim. = Cosine Similarity Threshold.}
\end{table*}

Table~\ref{tab:csthreshold} reports HPE-CogVLM performance under different cosine similarity thresholds. We observe that lower thresholds (e.g., 0.7, 0.8) introduce more layers from the HPE-oriented model, which improves HPE accuracy but leads to blended BBox outputs and noticeably lower detection accuracy. Conversely, very high thresholds (e.g., 0.98) preserve grounding well but slightly reduce HPE precision. 
Based on this analysis, we selected a threshold of 0.95 because it offers the best trade-off. It preserves high grounding accuracy across all RefCOCO datasets while maintaining strong HPE performance (MAE = 7.36 on CMU Panoptic). Importantly, the results also show that our layer-based merging method is not overly sensitive to the threshold: settings of 0.9, 0.95, and even 0.8 achieve consistently strong 
performance. This robustness indicates that the proposed winner-takes-all merging strategy effectively balances grounding preservation and HPE adaptation across a broad threshold range.
\\

\subsection{Performance of HPE-oriented CogVLM on HPE Task Only}

\begin{table}[t!]
\caption{The HPE-oriented CogVLM model exhibits the highest HPE performance within our framework. The best results are highlighted in bold.}
\centering
\resizebox{\linewidth}{!}{
\begin{tabular}{lcccccc}
\toprule
\textbf{Model} & \textbf{Epochs} & \textbf{$\text{Acc}_\text{test}$↑} & \textbf{$\text{MAE}_\text{test}$↓} & \textbf{$\text{MAE}_\text{train}$↓} & \textbf{$\text{E}_\text{angle}$↓} \\
\midrule
6DRepNet & 3 & - & 12.70 & 9.40 & - \\
6DRepNet & 6 & - & 12.76 & 8.80 & -\\
6DRepNet & 9 & -  & 11.44 & 7.90 & -\\
6DRepNet & 50 & -  & 11.37 & 2.91 & -\\
6DRepNet & 100 & -  & 11.4 & 2.23 & -\\
HPE-oriented CogVLM & 3 & 8.8\%  & 6.40 & - & 0.0092\%\\
HPE-oriented CogVLM & 6 & 12.6\% & 6.31 & - & 0\%\\
HPE-oriented CogVLM & 9 & 11.0\% & \textbf{6.24} & - & 0\%\\

\bottomrule
\end{tabular}
}

\label{tab:brute force}
\end{table}

In our framework, The HPE-oriented CogVLM from stage 2 stands out as the most effective model dedicated solely to the HPE task. Table \ref{tab:brute force} presents comparative performance results of 6DRepNet and the HPE-oriented CogVLM, both not accommodate BBox prediction capabilities, over similar training epochs. The low Refcoco test accuracy of HPE-oriented CogVLM is expected, given that no data rehearsal is implemented in this stage so that the VLM can focus on the HPE task learning. In terms of MAE metric, the HPE-oriented CogVLM displays a gradual decrease in MAE from 6.4 at 3 epochs to 6.24 at 9 epochs. When compared our model with 6DRepNet, in the same epoch, our MAE shows much lower numbers than 6DRepNet. For example, in epoch 9, MAE of HPE-oriented CogVLM is 6.24 which is 45.5\% lower than 6DRepNet. After extending the 6DRepNet training to 100 epochs, while its training MAE decreases from 9.40 to 2.23, the MAE on the CMU dataset does not improve, remaining stable around 11.4. This indicates that the model is over-fitting to the training dataset, with no enhancement in cross-dataset inference performance. This difference emphasizes the superior performance of VLM than traditional CNN-based models.

\subsection{Visualization of Cross Attention Maps Supervised by Designed Prompts}

\begin{figure}[h]
    \centering
    \includegraphics[scale = 0.22]{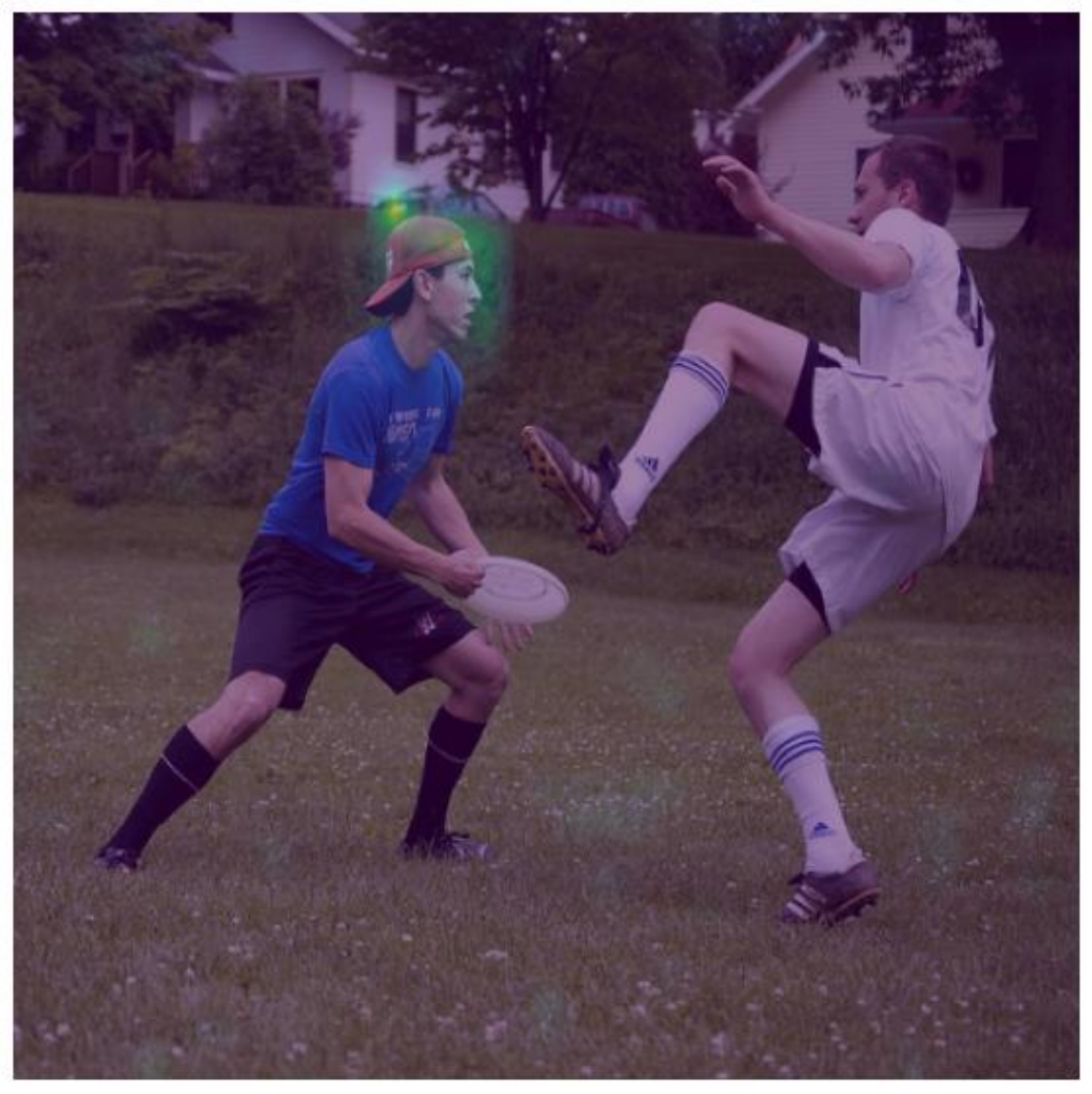}
    \includegraphics[scale = 0.22]{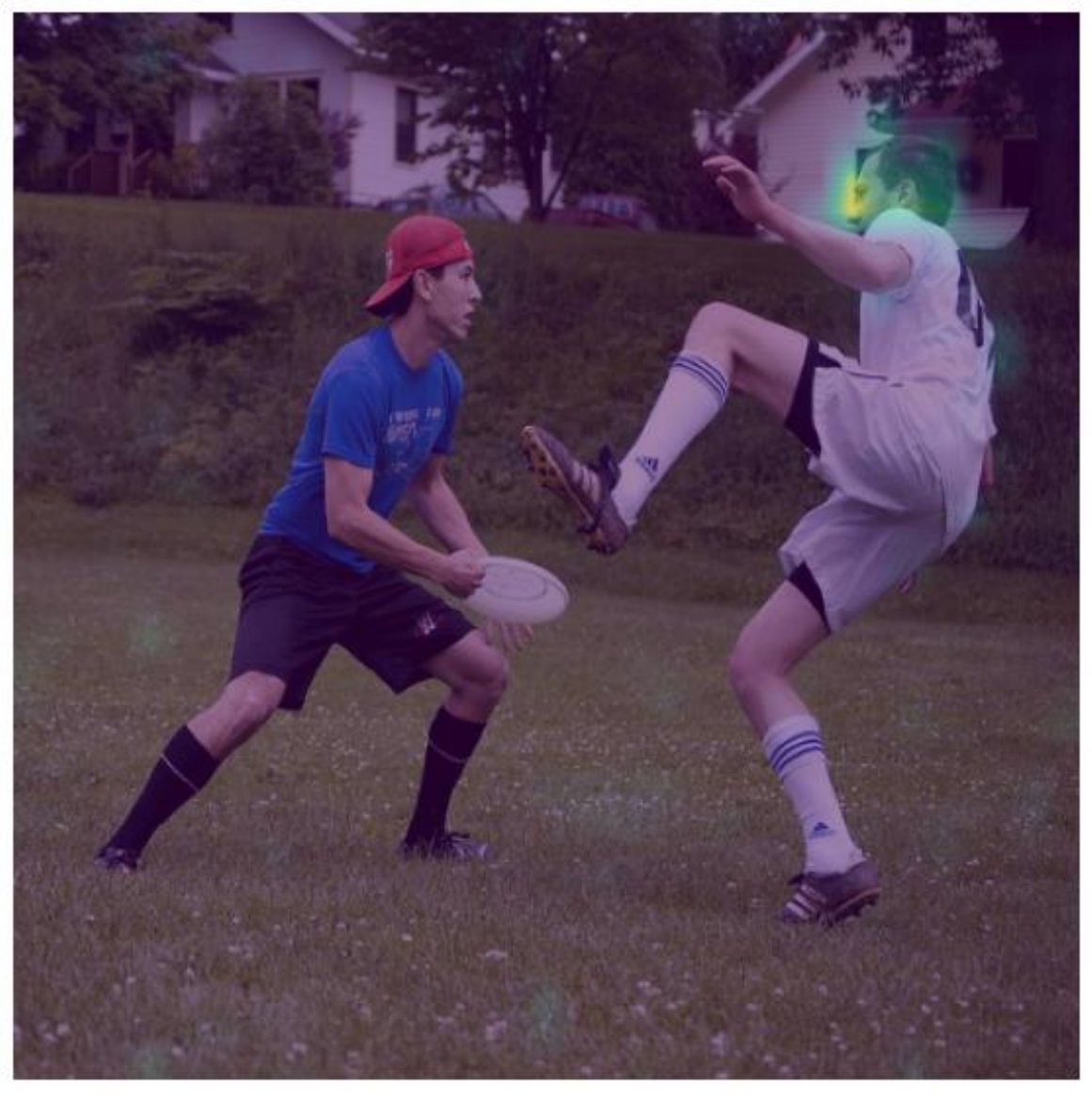}
    \caption{The visualization displays cross attention maps generated in response to our custom prompts. The left image shows the attention map associated with the prompt "What is the head yaw pitch roll inside the bounding box [[335,179,445,332]]?" (BBox for the person on the left), and the right image corresponds to the prompt "What is the head yaw pitch roll inside the bounding box [[775,105,893,261]]?" (BBox for the person on the right).}
    \label{fig:cross attn}
\end{figure}
Figure \ref{fig:cross attn} provides a visual representation of cross attention maps created in response to specific prompts when multiple people are present in the image. The left image highlights the model's response to the prompt of HPE task within the BBox [[335,179,445,332]]. It effectively focuses on the head of the person on the left. The right image similarly demonstrates the model's precision in targeting the head of the person on the right within the BBox [[775,105,893,261]]. These visualizations confirm the model’s accuracy in localizing attention within specified areas, demonstrating strong capabilities in spatial awareness. Moreover, they illustrate that CogVLM is capable not only of generating BBox outputs in response to queries but also of interpreting and responding to BBoxes specified within the prompts.

\begin{table*}[t]
\small
\centering
\caption{Comparison of output-parsing and constrained-decoding strategies vs our solution.}

\begin{tabular}{l p{4.2cm} c c}
\toprule
Method & Description & \textbf{$\text{E}_\text{angle}$↓} & MAE $\downarrow$ \\
\midrule
TA Merge & Baseline & 68.9\% & 7.72 \\
\midrule
\multicolumn{4}{l}{\textit{TA Merge + output parsing / constrained decoding:}} \\
\quad Regex-Loose   & Extract first three numeric values found in output & 21.0\% & 12.42 \\
\quad Regex-Strict  & Accept only outputs with exactly three values in [0,360] & 67.7\% & 7.86 \\
\quad Logit Constraint & Constrain vocabulary to digits and symbols & 49.1\% & 8.71 \\
\midrule
\textbf{Layer Merge (ours)} & Layer-wise merging strategy & \textbf{0.052\%} & \textbf{7.36} \\
\bottomrule
\end{tabular}

\label{tab:constraint_summary}
\end{table*}

\subsection{Evaluation of Output Format Enforcement Approaches}

To investigate whether the invalid outputs observed in TA-merged models 
(68.9\% as shown in Table~\ref{tab:model_performance}) could be addressed through output-level 
interventions, we evaluated three experiments based on relaxed output parsing and constrained decoding.

\textbf{Regex-Loose} extracts the first three numeric values from any output, 
ignoring count, range, and formatting constraints. For example, given the 
output ``the angle is \{11, 211, 312, 71, 21\}'', this method treats it as 
valid and uses 11, 211, and 312 as the yaw, pitch, and roll predictions.

\textbf{Regex-Strict} accepts only outputs containing exactly three values 
within [0,360] range, while relaxing other formatting requirements.

\textbf{Logit Constraint} applies constrained decoding~\cite{Hokamp2017LexicallyCD,Post2018FastLC} 
by masking non-numeric and non-structural tokens (setting their logits to 
$-\infty$) during generation, restricting the vocabulary to digits, symbols, 
and a minimal set of task-specific tokens.

Table~\ref{tab:constraint_summary} compares output format enforcement approaches with our layer-based merging strategy. Across all three methods, we observe a consistent pattern: output parsing (Regex-Loose, Regex-Strict) or token-level decoding constraints (Logit Constraint) reduces certain invalid outputs but introduces significant accuracy trade-offs.

Regex-Loose achieves the lowest invalid rate (\textbf{$\text{E}_\text{angle}$}=21.0\%) by greatly loosening the regex conditions described above, but this permissiveness results in a 61\% increase in MAE (from 7.72 to 12.42). Regex-Strict, in contrast, largely maintains accuracy but fails to meaningfully reduce invalid
outputs (\textbf{$\text{E}_\text{angle}$} = 67.7\%). Its requirement of exactly
three values in [0,360] remains strict enough to reject most malformed responses.
For instance, output with wrong counts \{12, 45, 78, 90]]) will be rejected. Only edge cases with exactly three valid-range numbers but minor formatting deviations (e.g., wrong close brackets like \{001, 359, 002]]) pass validation, explaining why \textbf{$\text{E}_\text{angle}$} improvement is minimal (1.2\%). This result illustrates another limitation of output parsing: the method is highly sensitive to how aggressively the validity criteria are relaxed. Logit Constraint lies between the two (49.1\% \textbf{$\text{E}_\text{angle}$}, 8.71 MAE), but still fails to produce reliable structured numeric outputs; this occurs because CogVLM’s constrained decoding inherently relies on static invalid-slice masking, which filters tokens independently and cannot enforce sequence-level structure (e.g., producing exactly three numbers rather than four or more).

Importantly, methods that relax validity criteria to accept more outputs including Regex-Loose and Logit Constraint both exhibit MAE degradation. This suggests that the invalid outputs originate from representation-level limitations in the TA-merged model’s HPE capability rather than from superficial formatting errors. Forcing them into valid format does not improve, and often degrades accuracy. In contrast, our layer-based merging addresses representation-level interference at its source, achieving both stable format (0.052\% invalid) and 
improved accuracy (7.36 MAE) without any post-hoc interventions.\\

\begin{table*}[t]
\small
\centering
\caption{Invalid output breakdown by scene composition.}

\begin{tabular}{lccccc}
\toprule
\multirow{2}{*}{\textbf{Scenario}} &
\multicolumn{2}{c}{\textbf{TA Merge}} &
\multicolumn{2}{c}{\textbf{Layer-Based Merge}} \\
\cmidrule(lr){2-3} \cmidrule(lr){4-5}
& \textbf{Invalid Outputs} & \textbf{$\text{E}_{\text{ang}}$↓} 
& \textbf{Invalid Outputs} & \textbf{$\text{E}_{\text{ang}}$↓} \\
\midrule
Single-Frontal & 3057 & 86.55\% & 2 & 0.057\% \\
Single-Back    & 3223 & 85.74\% & 3 & 0.080\% \\
Multi-Frontal  & 8250 & 63.63\% & 6 & 0.046\% \\
Multi-Back     & 8026 & 64.30\% & 6 & 0.048\% \\
\bottomrule
\end{tabular}

\label{tab:breakup}
\end{table*}

\subsection{Invalid Output Patterns by Scene Composition}

To analyze when and why invalid outputs occur, we compare error patterns by input complexity in Table~\ref{tab:breakup}. For the TA merged model, single-head cases exhibit significantly higher invalid ratios ($\sim$86\%) than multi-head cases ($\sim$64\%). This occurs because our HPE training data (Agora) contains only multi-person scenes. As a result, single-person images fall outside the TA model’s training distribution and more easily trigger interference from the original captioning/grounding behavior. Notably, this pattern does not change the main conclusion. Even in the most favorable multi-head scenes, the invalid ratio still exceeds 60\%. This shows that the issue arises from the TA merged capability itself rather than the scene type.

Conversely, HPE-CogVLM maintains consistently near-zero invalid rates ($<$0.1\%) across all categories. This comparison empirically proves that our layer-based merging strategy successfully disentangles task representations and prevents the format collapse. \\

\begin{figure*}[t]
    \centering
    \includegraphics[width=\linewidth]{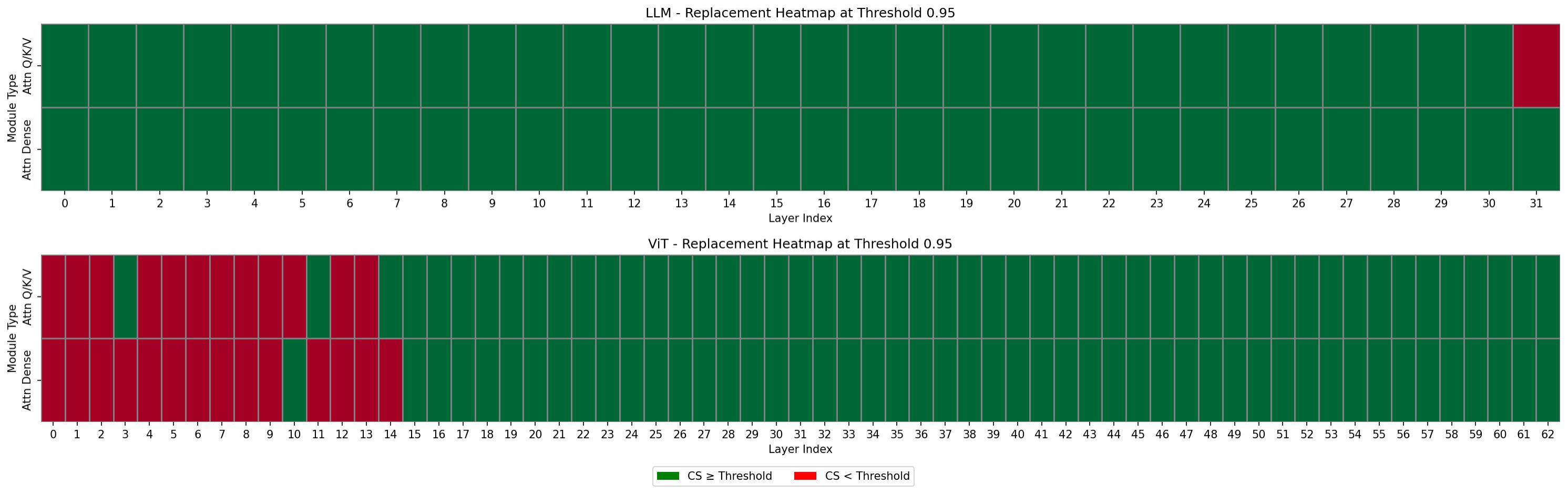}
    \caption{Layer-wise replacement pattern under cosine similarity $threshold=0.95$. Green cells indicate layers selected from the HPE-oriented model, while red cells indicate layers preserved from the original grounding model.}
    \label{fig:replacement_heatmap}
\end{figure*}

\subsection{Rotation Metrics and Pose-Range Analysis}

\begin{table}[t] 
    \centering 
    \caption{Per-angle MAE and geodesic rotation error under front/back pose ranges (yaw in $[-180,180]$; front: $|yaw|\le90$, back: $|yaw|>90$).}
    \begin{tabular}{llcccc} \hline Method & Range & Yaw $\downarrow$ & Pitch $\downarrow$ & Roll $\downarrow$& Geodesic $\downarrow$ \\ \hline 6DRepNet & Front &12.95 &7.99 &7.91 & 19.15 \\ HPE-CogVLM & Front & 8.77 & 6.60 &5.93 & 13.46\\ \hline 6DRepNet & Back & 15.75& 9.63& 10.23 & 26.18 \\ HPE-CogVLM & Back & 9.10 &6.39 & 7.39 & 15.13\\ \hline \end{tabular} 
    \label{tab:frontback_ge} 
\end{table}

In the experiment, we report both per-angle MAE (yaw, pitch, roll) and geodesic rotation error under front ($|yaw|\le90$) and back ($|yaw|>90$) pose ranges (Table~\ref{tab:frontback_ge}). The geodesic rotation error~\cite{COBO2024110263} is computed as the angular distance between the predicted and ground-truth rotation matrices derived from the corresponding yaw–pitch–roll angles. In the front-facing regime, HPE-CogVLM consistently reduces yaw, pitch, roll MAE and lowers geodesic error from 19.15 to 13.46 compared to 6DRepNet. In the more challenging back-facing regime, improvements remain substantial, with geodesic error decreasing from 26.18 to 15.13, alongside consistent per-angle MAE reductions. Furthermore, the performance gap between front and back-facing poses is substantially smaller for HPE-CogVLM (yaw MAE difference: $0.33$) compared to 6DRepNet ($2.80$), indicating that our full-scene approach is more robust to back-facing orientations than cropped-head CNN baselines. Overall, the reported 31.5\% reduction in MAE and 36.9\% reduction in geodesic rotation error reflects robust improvements across both evaluation metrics.

\begin{table*}[h]
\centering
\caption{Prompt formulation sensitivity analysis.}
\begin{tabular}{p{8cm}cc}
\hline
Prompt Formulation & MAE $\downarrow$ & $E_{angle}$ $\downarrow$ \\
\hline
What is the head yaw pitch roll inside the bounding box [[...]]? (Training prompt) & 7.36 & 0.052\% \\

For the bbox [[...]], what is the head yaw pitch roll? & 7.40 & 0.058\% \\

Please give the head pose yaw, pitch, roll for the bounding box [[...]]. & 7.37 & 0.33\% \\

Predict head pose yaw, pitch, roll in [[...]]. & 7.44 & 0.45\% \\
\hline
\end{tabular}
\label{prompt}
\end{table*}

\subsection{Layer-wise Analysis of Winner-Takes-All Merging}

To better understand the behavior of the proposed layer-wise winner-takes-all merging strategy, we analyze the replacement pattern under the selected cosine similarity threshold $\tau = 0.95$. Figure~\ref{fig:replacement_heatmap} visualizes the per-layer selection results. A clear structural pattern emerges rather than arbitrary layer swapping.

\textbf{LLM Backbone.}
The language backbone is almost entirely selected from the HPE-oriented model, with only the final QKV projection layer preserved from the original grounding model. This indicates that HPE fine-tuning introduces minimal disruption to language-generation components while maintaining structured numeric decoding stability.

\textbf{Vision Encoder (ViT).}
In contrast, early ViT attention layers (0--14) show lower cosine similarity, indicating stronger task-specific divergence after HPE fine-tuning. These layers are therefore preserved from the original grounding model according to our threshold and safeguard design. HPE adaptation is instead absorbed by deeper ViT layers (15--62), which show high similarity and are safely replaced by the HPE-oriented model. The subsequent Stage 4 joint fine-tuning further harmonizes both objectives.

In summary, the heatmap shows that most LLM layers and deeper ViT layers are selected from the HPE-oriented model, while early ViT layers remain from the original grounding model.\\

\subsection{Prompt formulation sensitivity analysis}

In this experiment, we compare four HPE results with different prompts. The results show MAE remains stable across all variants (7.36--7.44), indicating that HPE accuracy is robust to prompt phrasing and once the task intent is recognizable, HPE accuracy is reliably maintained. This demonstrates that the performance gain is associated with the architectural integration rather than prompt-specific overfitting. $E_{angle}$ shows sensitivity to prompt variation. This is expected, as the output format convention (e.g., \{yaw, pitch, roll\}) is learned in conjunction with the training prompt. When inference prompts deviate from the training distribution, the model may occasionally produce outputs in alternative formats. Nevertheless, even under unseen prompt formulations, $E_{angle}$ remains below 0.45\%, which is still substantially lower than TA merging (68.9\%), confirming that the structural advantage of our layer-based merging is robust to prompt variation. Notably, prompts that closely resemble the training format (e.g., prompt 2) maintain near-identical $E_{angle}$, while more syntactically distant formulations (prompts 3--4) show higher invalid rates, consistent with the model's sensitivity to output format triggers learned during training.\\

\section{Conclusions} \label{conclude}
In this paper, we present a framework that successfully enhances the HPE task by leveraging the visual grounding capabilities of CogVLM. Through a novel merging approach that utilizes a high cosine similarity threshold and a “winner-takes-all” layer selection strategy, we effectively integrate HPE capabilities into the model while preserving the original BBox knowledge. This method not only improves prediction accuracy but also addresses the challenges of blended invalid response formats. Additionally, we mitigate catastrophic forgetting by optimizing the rehearsal ratio. Our experimental results demonstrate that HPE-CogVLM achieves a substantial 31.5\% reduction in MAE compared to the current CNN-based state-of-the-art in cross-dataset evaluations. Furthermore, HPE-CogVLM consistently outperforms directly LoRA fine-tuned model and task arithmetic-based merging model in all HPE metrics, establishing it as a robust approach for improving and stabilizing structured numerical prediction within grounding-oriented VLMs.

\section{Limitations and Future Work}
\textbf{Limitations.} While HPE-CogVLM alleviates reliance on tight head crops and shows strong robustness in cluttered, multi-person scenes, it shares fundamental limitations with all HPE models. When BBox grounding is inaccurate or when the head region suffers from visual degradation, pose estimation becomes unreliable. Such degradation includes extreme lighting conditions (e.g., backlighting or low-light environments), heavy occlusion of key facial regions (by hair, masks, or others’ limbs), and rare extreme orientations (roll or pitch beyond ±90°). These cases offer insufficient geometric evidence and are underrepresented
in existing training datasets, which limits performance. Nevertheless, by operating on full images with contextual grounding, our approach remains more resilient than traditional cropped-head methods.

\textbf{Application Directions.} Despite these limitations, HPE-CogVLM shows strong potential for real-world deployment by operating on detected heads within full-scene context, unlike traditional methods limited to cropped faces. In driver monitoring, context cues such as steering wheels or seat layouts help maintain accuracy even when drivers turn fully away. In retail analytics, the framework enables robust HPE in crowded stores, allowing inference of customer focus even when they are back-facing the camera.  In human–robot interaction, it unifies full-scene pose estimation with language grounding, enabling robots to answer natural language queries such as “Where is the user looking?” while estimating head orientation. 

\textbf{Future Work.} This work shows that layer-based architectural separation can improve and stabilize structured numerical outputs in VLMs. We use yaw–pitch–roll HPE as a representative case. The same idea may extend to other geometric tasks such as 6-DoF pose estimation~\cite{xiang2018posecnnconvolutionalneuralnetwork} or object-level depth estimation~\cite{mousavian20173dboundingboxestimation}. However, extending to these domains would require adapting the output tokenization strategy to represent spatial translations or depth metrics, and utilizing task-specific datasets with distinct evaluation protocols.
\\

\end{document}